%% file: main.tex
\documentclass{article}

 \usepackage[preprint]{neurips_2026}

\usepackage{microtype}
\usepackage{graphicx}
\usepackage{subcaption}
\usepackage{booktabs}
\usepackage{amsmath}
\usepackage{amssymb}
\usepackage{mathtools}
\usepackage{amsthm}
\usepackage{enumitem}
\usepackage{caption}
\usepackage{wrapfig}

\usepackage[utf8]{inputenc} % allow utf-8 input
\usepackage[T1]{fontenc}    % use 8-bit T1 fonts
\usepackage{hyperref}       % hyperlinks
\usepackage{url}            % simple URL typesetting
\usepackage{booktabs}       % professional-quality tables
\usepackage{amsfonts}       % blackboard math symbols
\usepackage{nicefrac}       % compact symbols for 1/2, etc.
\usepackage{microtype}      % microtypography
\usepackage{xcolor}         % colors

\title{Two-Bridge: Exclusive Objectives and Extended Horizon StarCraft II Benchmark}

\author{%
  Sourav Panda\textsuperscript{1}
  \thanks{
  \textsuperscript{1}Pennsylvania State University, PA, USA.\\
  Emails: \texttt{\{sbp5911,tanmay.ambadkar,verma,dodge\}@psu.edu}.
  Correspondence to: Sourav Panda, \texttt{sbp5911@psu.edu}.
  }
  \And
  Tanmay Ambadkar\textsuperscript{1}
  \And
  Shreyash Kale\textsuperscript{1}
  \AND
  Abhinav Verma\textsuperscript{1}
  \And
  Jonathan Dodge\textsuperscript{1}
}

\begin{document}

\maketitle

\begin{abstract}
\input{documentBody/0-Abstract}
\end{abstract}

%\clearpage
\input{documentBody/1-introduction}
%\clearpage
\input{documentBody/2-tbmEnv}
%\clearpage
\input{documentBody/3-results}
%\clearpage
\input{documentBody/4-limAndFut}
%\clearpage
\input{documentBody/5-conclusion}

\clearpage
\bibliography{main}
\bibliographystyle{unsrt}

%%%%%%%%%%%%%%%%%%%%%%%%%%%%%%%%%%%%%%%%%%%%%%%%%%%%%%%%%%%%
\clearpage
\appendix

\input{documentBody/Appendix/A-Code}
\input{documentBody/Appendix/B-mainResults}
\clearpage
\input{documentBody/Appendix/C.1-pilot1}
\input{documentBody/Appendix/C.2-pilot23}
\input{documentBody/Appendix/D-Trigger}

%%%%%%%%%%%%%%%%%%%%%%%%%%%%%%%%%%%%%%%%%%%%%%%%%%%%%%%%%%%%

\newpage

\end{document}

%% file: documentBody/0-Abstract.tex
%``Dr. Burnett's Formula'': 
% 1. What's the problem
% 2. Why is the problem a problem
% 3. What did we do about it?
% 4. What did that do for the world OR what are 1-2 interesting things we found out?
The research community lacks a middle ground between StarCraft II’s full game and its mini-games. 
The full-game's sprawling state-action space renders reward signals sparse and noisy, but in mini-games simple agents saturate performance.
This complexity gap hinders steady curriculum design and prevents researchers from experimenting with modern Reinforcement Learning algorithms in RTS environments under realistic compute budgets.
To fill this gap, we present the Two-Bridge Map Suite, the first entry in an open-source benchmark series we purposely engineered as an intermediate benchmark to sit between these extremes.
By disabling economy mechanics such as resource collection, base building, and fog-of-war, the environment isolates two core tactical skills: long-range navigation and micro-combat.
Preliminary experiments show that agents learn coherent maneuvering and engagement behaviors without imposing full-game computational costs.
Two-Bridge is released as a lightweight, Gym-compatible wrapper on top of PySC2, with maps, wrappers, and reference scripts fully open-sourced to encourage broad adoption as a standard benchmark.

%% file: documentBody/1-introduction.tex
\section{Introduction}

%%%%%%%%%%%%%%%%%%%%%%%%%%%%%%%%
\input{figure/benchmarkGap}
%%%%%%%%%%%%%%%%%%%%%%%%%%%%%%%%

Games have emerged as a central testbed for RL research due to their well-defined dynamics and controllable simulation settings~\cite{shao2019survey, szita2012reinforcement, lanctot2019openspiel, lample2017playing, erev1998predicting}.
StarCraft~II (SC2) is widely recognized as a premier environment for RL research due to its combination of long-horizon planning, multi-agent coordination, partial observability, and real-time control~\cite{Vinyals2017, ontanon2013survey, wang2021scc, liu2022efficient, li2024multiagent, pang2019reinforcement, mathieu2021starcraft, xu2020hierarchical, sun2018tstarbots}. 
The game naturally decomposes into macromanagement, involving high-level strategy, and micromanagement, involving fine-grained unit-control. 
This hierarchical structure makes SC2 a flexible platform for studying RL problems ranging from long-term planning to tactical decision-making.
Moreover, SC2 allows researchers to isolate different aspects of the problem through custom scenarios, enabling targeted benchmarks while retaining core domain complexity.

\textbf{Limitations of existing SC2 benchmarks.}
Prior SC2 benchmarks span full-game, mini-game, custom-scenario, and SMAC settings, each balancing domain richness and usability differently.
AlphaStar~\cite{vinyals2019grandmaster} achieved Grandmaster-level performance in full-game SC2 using large-scale league-based training, imitation learning from human replays, and specialized neural architectures. 
However, reproducing such systems remains infeasible for most researchers due to extreme compute requirements, complex training pipelines, proprietary components, and the sheer scale of interaction required, estimated to exceed \textit{trillion} environment steps~\cite{dodge2021billion}. 
Later approaches reduce complexity through scripting, modular architectures, macro-actions, or offline replay-based training~\cite{lee2018modulararchitecturestarcraftii,sun2018tstarbotsdefeatingcheatinglevel,pang2019reinforcementlearningfulllengthgame,Liu2022OnER,mathieu2023alphastarunpluggedlargescaleoffline}, but often rely on hand-crafted components, large replay corpora, or restricted assumptions. 
Thus, full-game remains impractical as a broadly \emph{usable} research testbed under limited compute budgets.

SMAC~\cite{samvelyan2019starcraft} is a widely used cooperative MARL benchmark~\cite{wang2020qplex, mahajan2019maven, wang2020roma, rashid2020monotonic, rashid2020weighted, wu2025learning, yu2022surprising, wang2020rode}. 
However, it is limited to micro-level combat, whereas full-game SC2 also involves macromanagement, navigation, exploration, and higher-level strategic reasoning. 
By isolating combat, SMAC reduces the problem to coordination under fixed combat settings. 
It is therefore inherently single-objective, with success defined only by defeating enemy units, unlike the full game where objectives can be multiple and sometimes mutually exclusive. 
This also makes evaluation largely win-rate based, which can hide behavioral diversity among policies that achieve the same outcome~\cite{panda2025position}; 
for example, two agents may reach similar win rates through focus fire, distributed targeting, kiting, or terrain-based positioning.
SMAC also relies on centralized training with decentralized execution (CTDE), evaluating policies under one  training--deployment paradigm. 
SMAC uses only the raw SC2 API, where agents receive unit-level numerical features such as health, distance, and relative positions. 
Although relative positions encode some spatial information with respect to the map, they do not fully capture the graphical and perceptual structure of the environment. 
Thus, the spatial reasoning problem is partly reduced to feature-based optimization, limiting the extent to which agents must learn representations grounded in the broader spatial dynamics of SC2.
SMACv2~\cite{ellis2023smacv2} extends SMAC with additional stochasticity and scenario variation. 
However, recent work shows that SMAC is near saturation, with many algorithms achieving near-perfect win rates on most scenarios~\cite{hu2021rethinking, gorsane2022towards, de2020independent}.

SC2 mini-games~\cite{Vinyals2017} provide lightweight tasks for studying isolated RL skills under modest computational budgets~\cite{zambaldi2018relationaldeepreinforcementlearning,8959866,xu2020hierarchicalreinforcementlearningstarcraft}. 
However, they are narrowly scoped and typically focus on single skills or objectives. 
Custom SC2 scenarios have similarly been developed for specific learning objectivess~\cite{shao2018starcraft,song2020spellcaster,huang2021adversary,waytowich2019narration,khanna2022finding}. 
While these scenarios demonstrate the flexibility of SC2, they are usually designed around individual research questions rather than standardized benchmark suites for systematic comparison.
Figure~\ref{fig:Benchmark_Gap} highlights the compact spatial scale of mini-games and SMAC relative to full-game SC2.
In SMAC, closely initialized units make interactions immediately accessible, shortening the decision horizon and reducing the need for search, navigation, or delayed objective discovery. 
Even SMACv2's extended partial observability is introduced through stochastic visibility masking rather than emerging naturally from larger spatial separation.
This leaves a missing middle in the SC2 benchmark landscape: 
an environment that is more strategically structured than mini-games and SMAC-style combat tasks, but still far more usable than full-game SC2.

\textbf{Our Contribution:} 
We introduce the \emph{Two-Bridge Map Suite}, an intermediate SC2 benchmark designed to bridge the gap between constrained micro-scenarios and full-game SC2.
Two-Bridge preserves meaningful tactical decision-making while remaining reproducible and accessible under realistic compute budgets.
Our contributions are:
(i) \textbf{an SC2 benchmark suite} that isolates tactical behavior while retaining nontrivial decision horizons; and
(ii) \textbf{a lightweight, Gym-compatible interaction layer} built on PySC2, enabling plug-and-play training with standard RL algorithms.
Two-Bridge is not intended as the only possible intermediate benchmark, but as one instantiation of a broader class of accessible SC2 environments for progressive RL research.
While some variants are solvable with existing methods, others remain challenging for standard RL baselines, making the suite useful for both controlled evaluation and future algorithmic development.
Its compositional design enables a \emph{choose-your-own-adventure} style of evaluation, allowing researchers to select task variants that align with their investigative goals and available computational resources.

%% file: figure/benchmarkGap.tex
\begin{figure}
    \centering
    \includegraphics[width=\textwidth]{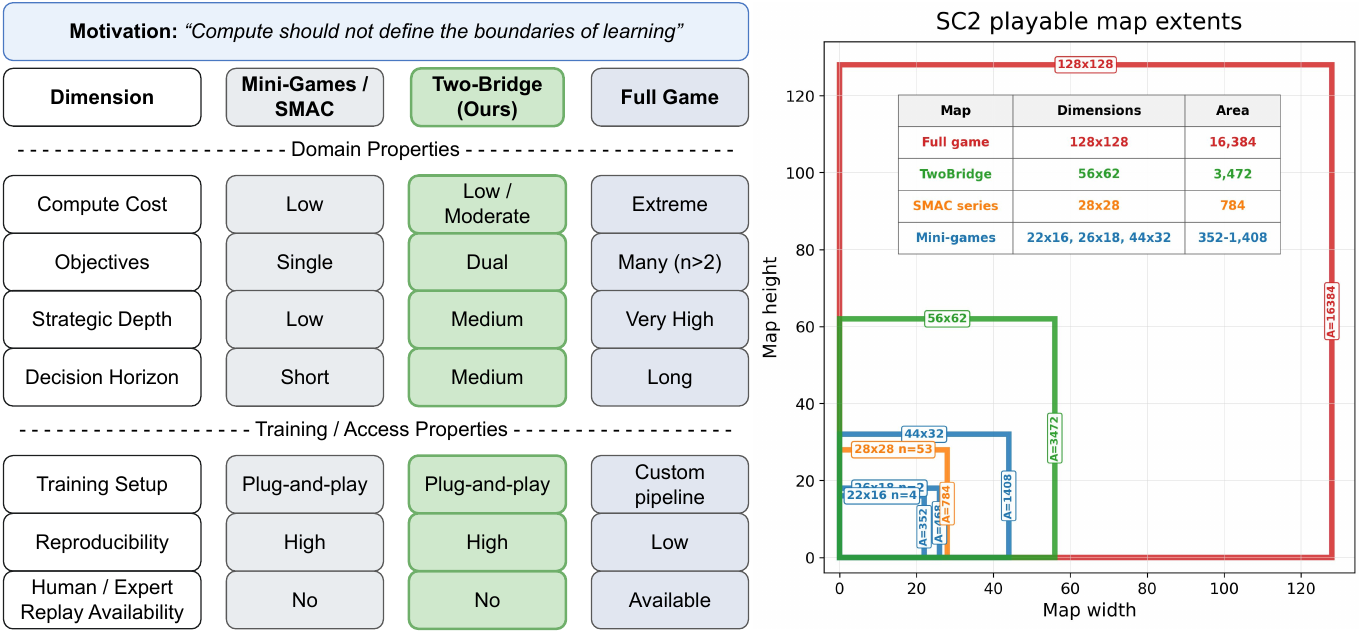}
    \caption{The benchmark gap in StarCraft II.
    \textbf{Left:} comparison of domain and access properties across SC2 benchmark families.
    \textbf{Right:} comparison of playable map extents.
    Two-Bridge targets an intermediate setting between constrained mini-games/SMAC scenarios and full-game SC2, introducing intermediate decision horizons and mutually exclusive objectives while remaining interactive, replay-free, and reproducible under standard compute budgets.}
    \label{fig:Benchmark_Gap}
\end{figure}

%% file: documentBody/2-tbmEnv.tex
\section{Two Bridge Map Suite}
\begin{wrapfigure}{r}{0.38\textwidth}
    \vspace{-1.2em}
    \centering
    \includegraphics[width=\linewidth]{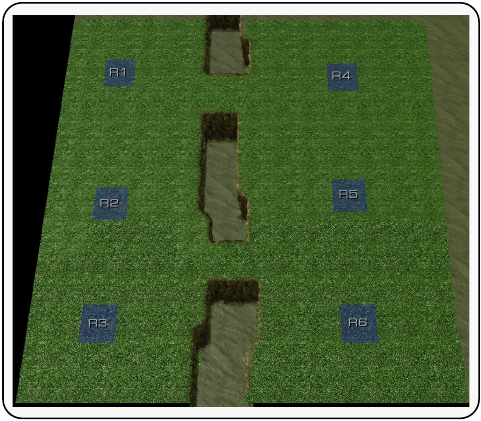}
    \caption{\textbf{Two-Bridge map layout.}}
    \label{fig:twoBridgeMap}
    \vspace{-1.0em}
\end{wrapfigure}

We designed the Two-Bridge Map Suite from scratch using the SC2 Editor, with editor-level implementation details deferred to Appendix~\ref{triggers}.
Figure~\ref{fig:twoBridgeMap} shows the terrain layout: a vertical cliff divides the playable area into left and right regions, connected only by two bridges. 
Because units cannot traverse the cliff directly, movement is forced through these bridges, creating narrow choke points and a clear spatial divide between the two sides.
The map defines six spawn regions, R1--R3 on the left and R4--R6 on the right, shown as blue overlays in Figure~\ref{fig:twoBridgeMap}.
These regions determine where units and the beacon can spawn.
The beacon is a static, non-controllable entity that appears in an unoccupied spawn region and remains stationary throughout the episode.

A unit is a controllable entity defined by position, health, movement capability, and attack range.
Friendly and enemy units spawn simultaneously as cohesive groups in distinct regions, with no reinforcements or respawns.
To isolate tactical decision-making, all units are homogeneous Marines limited to basic movement and ranged attacks, with no special abilities, upgrades, or resource-dependent mechanics.
Friendly units are controlled by RL agents in either centralized single-agent or decentralized multi-agent formulations, enabling different algorithms and policy architectures within the same environment.
Enemy units use the built-in SC2 AI, which is reactive in this custom map: enemies remain stationary until provoked and attack opposing units within range, without higher-level strategic adaptation.
Fog-of-war is disabled, making the environment fully observable and removing uncertainty from hidden-information settings~\cite{sokota2025superhuman, Perolat}.

\subsection{Challenges Posed by the Two-Bridge Benchmark}
Each Two-Bridge episode instantiates four core challenges: randomized initial conditions, mutually exclusive objectives, an extended search-and-execution horizon, and diagnostic terminal outcomes. 

\begin{wrapfigure}{r}{0.38\textwidth}
    \vspace{-1.2em}
    \centering
    \includegraphics[width=\linewidth]{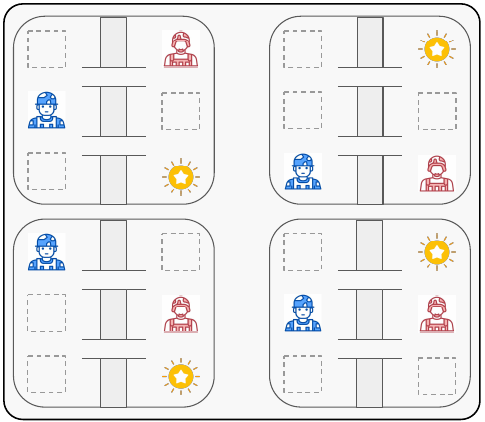}
        \caption{\textbf{Randomized Initial conditions.}}
        \label{fig:random_spawn}
    \vspace{-1.0em}
\end{wrapfigure}

\textbf{Randomized Initial Conditions:}
At the start of each episode, friendly units, enemy units, and the beacon spawn in predefined regions R1--R6.
Friendly and enemy units spawn as cohesive groups in distinct regions, while the beacon appears in an unoccupied region.
This spatial placement of unit groups and beacon varies across episodes, resulting in different initial configurations (Figure~\ref{fig:random_spawn});
blue icons denote friendly units, red icons denote enemy units, yellow stars denote beacon objectives, and dashed boxes denote spawn regions.
Randomization prevents the task from reducing to a single fixed path and requires agents to generalize across varying starting conditions.
Scenario configurations are sampled independently at each episode, not through any fixed or sequential ordering.

\textbf{Mutually Exclusive Objectives:}
Two-Bridge presents two tactical objectives within the same episode: \textbf{navigation} and \textbf{combat}.
Although this resembles a multi-objective RL setting, both objectives cannot be completed together; 
the agent must choose which objective to prioritize under the current spatial configuration.
This captures a simplified form of strategic choice in SC2, where agents often commit to one tactical opportunity at the expense of another.
\textbf{Navigation (Beacon Capture)} requires the agent to reach an \emph{unguarded} beacon spawned in an unoccupied region, requiring the agent to identify its location and plan and execute a feasible path.
\textbf{Combat (Eliminate Enemies)} requires the agent to navigate toward the enemy spawn region and perform fine-grained micromanagement during engagement.
By disabling economy, base construction, and technology progression, Two-Bridge focuses learning on movement, positioning, engagement, and objective selection.
By making navigation and combat mutually exclusive, Two-Bridge shifts the problem from simply executing available behaviors to deciding which objective to pursue.
This allows the benchmark to test preference selection, commitment, and tactical trade-offs while remaining substantially simpler than full-game SC2.
Figure~\ref{fig:random_spawn} illustrates both objectives present within the same map configuration.

\textbf{Search-and-Execution Horizon:}
In Two-Bridge, objectives are not immediately reachable from the initial spawn locations.
Because the playable area is larger than SC2 mini-games and SMAC-style scenarios, units must traverse the map before capturing the beacon or engaging enemies.
This introduces a spatial search phase before objective execution.
Unlike full-game SC2, the search is not driven by fog-of-war or hidden information; 
instead, agents must identify the relevant objective location, choose a route, and commit to movement through constrained bridge passages.
As a result, the challenge shifts from localized action execution to search-and-execution.
This extended horizon is important because early movement decisions can determine whether the agent reaches the beacon, encounters the enemy, takes an inefficient path, or fails to complete any objective within the episode limit.
In contrast to short-horizon micro-combat settings, where the primary challenge is immediate target selection or unit control, Two-Bridge requires agents to make decisions whose consequences unfold over a longer sequence of actions.
Although still far simpler than full-game SC2, this extended horizon preserves a similar source of difficulty by requiring agents to reason over delayed objective access rather than acting only on immediately available objectives.

\textbf{Diagnostic Terminal Outcomes:}
Each episode ends in one of four mutually exclusive outcomes: 
\textit{navigation win}, \textit{combat win}, \textit{combat loss}, or \textit{timeout loss}.
A navigation win occurs when a friendly unit captures the beacon, a combat win when enemies are eliminated, a combat loss when all friendly units are eliminated, and a timeout loss when no objective is completed within the episode limit of 5 minutes.
This structure is more diagnostic than aggregate win rate alone: 
agents with similar win rates may favor different objectives, while combat losses and timeouts expose failures in search, commitment, or tactical execution.

Together, these challenges shift the task from localized action execution to objective selection, route commitment, and long-horizon tactical control.
This makes Two-Bridge more structured than short-horizon mini-games and SMAC-style combat tasks, while remaining substantially more controlled and accessible than full-game SC2.

\input{figure/mapVariants}

\subsection{Map Variations and Strategic Diagnostics}
To evaluate agent behavior across tactical settings, we define variants along two orthogonal axes:
(i) \textbf{Unit Count Balance}, the relative number of friendly and enemy units; and
(ii) \textbf{Objective Proximity}, the spatial relationship between unit groups and the beacon.
\textbf{Unit Count Variants:}
Because all units are homogeneous Marines, numerical advantage determines combat difficulty.
We fix friendly units at five and vary enemy counts:
(i) \textbf{V1: Friendly Advantage (5F $>$ 3E):} friendly units outnumber enemies;
(ii) \textbf{V2: Balanced Forces (5F = 5E):} both sides have equal unit counts; and
(iii) \textbf{V3: Enemy Advantage (5F $<$ 8E):} enemies outnumber friendly units.
\textbf{Objective Proximity Variants:}
We vary the spatial arrangement of friendly units, enemies, and the beacon.
The left--right assignment is fixed but arbitrary; what matters is the relative proximity that changes the cost and risk of each objective.
(i) \textbf{Base -- Equidistant Objectives:} friendly units spawn on the left, while enemies and the beacon spawn on the right, making both objectives similarly distant.
(ii) \textbf{Combat-Proximal Variant:} friendly and enemy units spawn on the right, while the beacon spawns on the left, making combat closer.
(iii) \textbf{Navigation-Proximal Variant:} friendly units and the beacon spawn on the right, while enemies spawn on the left, making navigation closer.

\textbf{Benchmark Suite.}
Pairing the three unit-count settings with the three objective-proximity settings yields \textbf{nine configurations}.
Unit count controls combat risk, while objective proximity controls the cost of navigation versus engagement.
With randomized spawning, these variants test whether agents can generalize across initial conditions, choose between mutually exclusive objectives, and act over extended horizons.
Figure~\ref{fig:Map_Variants} summarizes the nine configurations and their strategic diagnostics.

\subsection{Training Recipe}
\label{sec:training_recipes}

PySC2 exposes several interaction channels for SC2, including screen features, minimap features, raw-unit observations, and a large action interface for full-game control. 
For Two-Bridge, exposing the full observation and action space is unnecessary and makes the learning problem harder in ways that are not central to the benchmark. 
We therefore use a reduced observation, action, and reward interface that preserves the tactical structure of the task while removing unused full-game complexity. 
This recipe was finalized after experimenting with several training configurations.
The details of these earlier recipes are not required for understanding the experiments in the main paper, so we defer them to Appendix~\ref{appendix:pilot1}~\ref{appendix:pilot2}~\ref{appendix:pilot3}.
The results reported in this paper use the recipe below, which we recommend as the standard interface for future algorithmic comparisons on Two-Bridge.

\begin{wrapfigure}{r}{0.38\textwidth}
    \vspace{-1.2em}
    \centering
    \includegraphics[width=\linewidth]{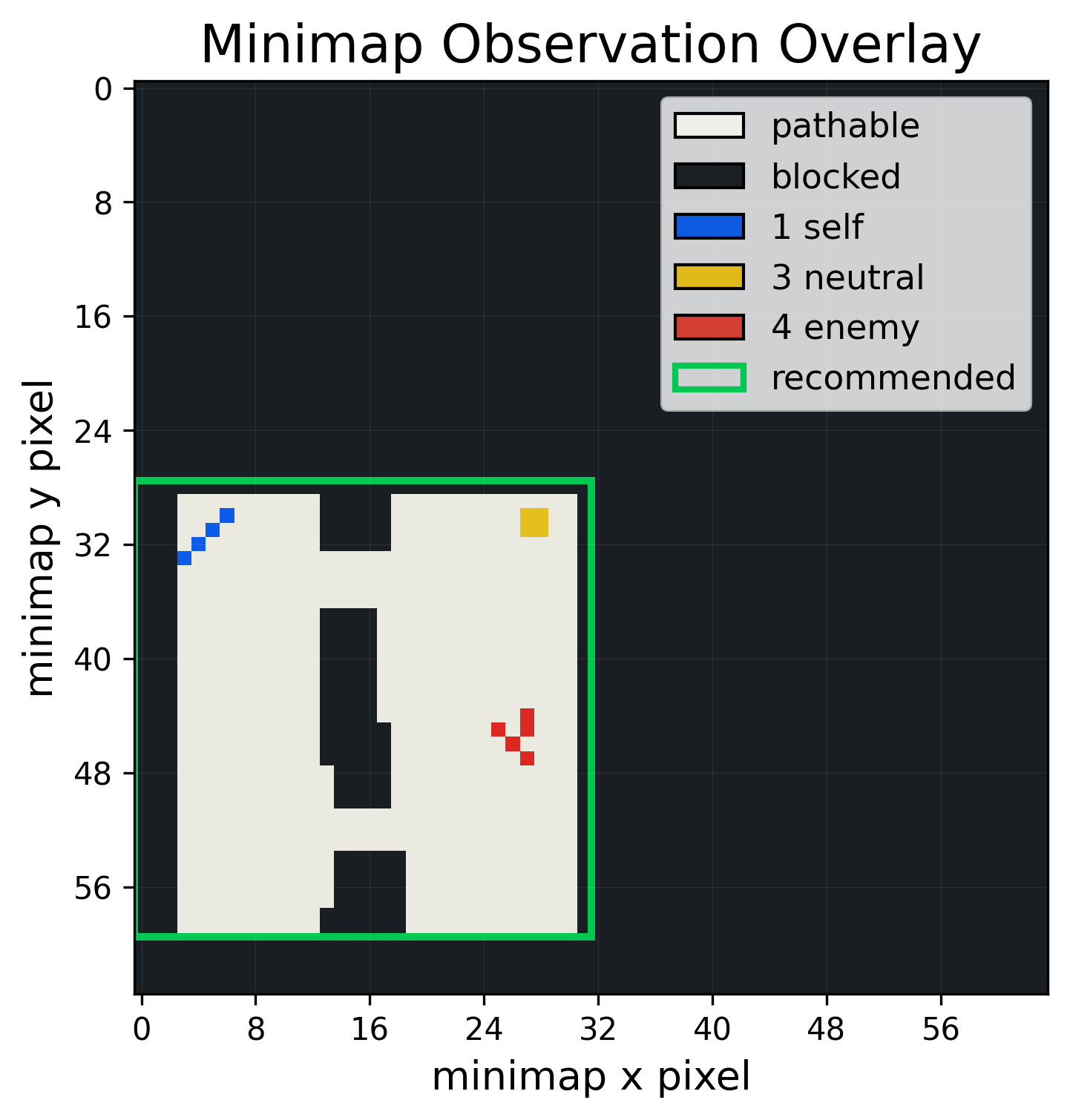}
    \caption{\textbf{Reduced minimap observation.}
    Cropped $32 \times 32$ minimap used as the spatial observation, with pathability and player-relative channels overlaid for visualization.}
    \label{fig:spatial_obs}
    \vspace{-1.0em}
\end{wrapfigure}

\textbf{State and Observations.}
At each timestep, the Two-Bridge observation has two components: a spatial component and a vector component. 
The spatial component is a reduced minimap describing the active Two-Bridge region. 
The original SC2 feature minimap is cropped to the playable bridge corridor, producing a $32 \times 32$ spatial input. 
In this setting, the minimap contains two channels: a pathability channel and a player-relative channel. 
The pathability channel identifies walkable and blocked terrain, while the player-relative channel provides a coarse spatial representation of allied, enemy, neutral, and empty regions on the minimap.
The spatial component is important because in Two-Bridge units must first move through a larger playable area, identify which side of the map contains the relevant objective, and navigate through bridge passages before they can complete either the navigation or combat objective. 
Since initial conditions are randomized, the policy cannot rely on a fixed route or a single memorized layout instance. 
The minimap therefore preserves the bridge structure and movement constraints needed for search and navigation, while avoiding unnecessary full-screen visual input.

The second component is a compact unit-status vector derived from raw-unit information. 
For each allied and enemy unit, this vector records whether the unit is alive and its current health. 
It also includes elapsed game time and the number of remaining enemies. 
We intentionally omit exact unit coordinates from the vector state, since providing them would reveal objective and unit locations directly, weakening the intended spatial search component of the benchmark.
These features are necessary for tracking the evolving combat state, as they determine unit-level actions at a given timestep, which we describe next. 
Together, the minimap and unit-status vector define the observation used across the algorithms evaluated in this paper.

\textbf{Action Space.}
Each controlled unit has three types of primitive actions: no-op, movement, and attack. 
The no-op action leaves the unit idle for the current step. 
Movement actions move the unit by a fixed step in one of eight directions: up, down, left, right, and the four diagonals. 
These movements are executed through raw SC2 unit commands, with target positions constrained to remain within the valid map region. 
Attack actions are defined over enemy units: each enemy slot in the scenario corresponds to one possible attack action.
For a scenario with $E$ enemy units, each allied unit has $9+E$ candidate actions: one no-op action, eight movement actions, and one attack action for each enemy slot. 
The number of actions varies with the number of enemies, while the action semantics remain consistent across all map variants.
Not all candidate actions are valid at every timestep. 
A dead allied unit cannot move or attack, and an attack action is valid only when the corresponding enemy is alive and within attack range, following the same attack-range constraint used in SMAC. 
We use action-availability masks to remove invalid commands before action selection, preventing policies from selecting impossible commands while keeping action interface consistent across algorithms.

\textbf{Reward Structure.}
Two-Bridge uses a shared team reward for all controlled units, aligned with its two mutually exclusive objectives: navigation and combat.
Both objectives use progress-based shaping and terminal feedback, but with different weights to reflect their learning difficulty.
\textbf{Navigation reward} has two components.
First, the agent receives distance-based shaping toward the beacon. 
At each timestep, the environment measures the distance from each living allied unit to the beacon and compares it with the previous timestep. 
If the allied units move closer to the beacon on average, the reward is positive; 
if they move farther away, the reward is negative. 
This provides intermediate feedback for route finding. 
Second, a high terminal reward is given when an allied unit reaches the beacon radius. 
This larger terminal bonus is used because the navigation objective is sparse.
\textbf{Combat reward} has four components.
First, the agent receives distance-based shaping toward the enemy group, computed using progress toward the centroid of the living enemy units. 
This encourages the allied units to search for and approach enemies. 
Second, the reward includes health-based feedback: reductions in enemy health are rewarded, while reductions in allied health are penalized. 
Third, the reward includes unit-level outcome feedback, rewarding enemy eliminations and penalizing allied losses. 
Finally, a positive terminal reward is given when all enemy units are eliminated. 
The combat terminal reward is lower than the navigation terminal reward because combat provides denser intermediate feedback than navigation.
Episodes terminate with one of four benchmark outcomes: navigation win, combat win, combat loss, or timeout loss. 
A navigation win receives a terminal reward of $+25$, a combat win receives $+10$, a combat loss receives $-10$, and a timeout loss receives $-15$.

%% file: figure/mapVariants.tex
\begin{figure*}[t]
    \centering
    \includegraphics[width=\textwidth]{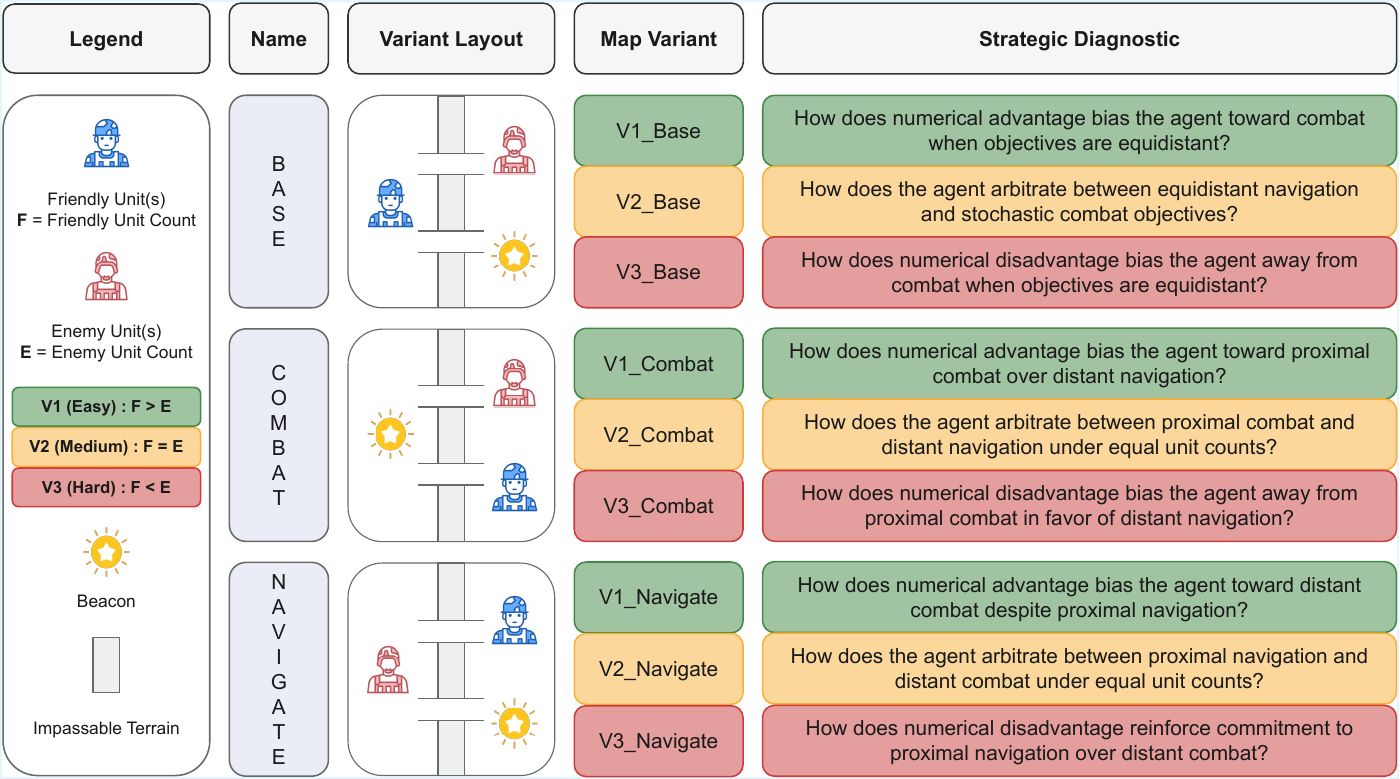}
    \caption{Two-Bridge map variants and strategic diagnostics.
    Each map is defined by the cross-product of a layout-induced proximity bias and a unit-count regime.
    \textbf{Takeaway:} Each resulting configuration, poses a strategic diagnostic that probes how agents arbitrate between objectives when structural affordances and numerical incentives are placed in tension.
    }
    \label{fig:Map_Variants}
\end{figure*}

%% file: documentBody/3-results.tex
\section{Experiments, Evaluation, and Results}

To validate the benchmark suite end-to-end, we evaluate it using both single-agent (MaskPPO~\cite{huang2020closer} from SB3) and multi-agent (MAPPO~\cite{yu2022surprising} and QMIX~\cite{rashid2020monotonic} from epymarl~\cite{papoudakis2021benchmarking}) settings. 
Our goal is to evaluate whether the benchmark exposes a learnable reward signal while eliciting meaningful strategic trade-offs between navigation and combat objectives.
Across all experiments, agent--environment interaction follows a standard pipeline in which the SC2 map is interfaced through PySC2, wrapped in a Gym-compatible API, and optimized by an RL agent. 
Agents are trained entirely from scratch without scripting, curriculum learning, demonstrations, or imitation signals, ensuring that successful behavior emerges solely through direct environment interaction.
Each agent is trained for 10M environment timesteps across all benchmark variants using 3 random seeds. 
The exact observation and action space configuration depends on the learning setup and is described in Appendix~\ref{appendix:training}. 
All experiments are conducted under modest consumer-grade compute constraints without distributed training or compute clusters; hardware and implementation details are provided in Appendix~\ref{hardware}.

\subsection{Evaluation Protocol}
We adopt a checkpoint-based evaluation protocol in the style of prior SC2 benchmarking methodology used in SMAC~\cite{samvelyan2019starcraft}, while adapting it to the Two-Bridge setting. 
During training, model checkpoints are saved at regular 50K timestep intervals. 
After training, each checkpoint is evaluated for 32 episodes using deterministic policy execution. 
For MaskPPO, this corresponds to deterministic action prediction. 
For MAPPO and QMIX, this corresponds to greedy action selection. 
In all cases, stochastic exploration is disabled during evaluation.

Because Two-Bridge contains two mutually exclusive success conditions, we define win rate as the fraction of evaluation episodes that terminate in either a navigation win or a combat win. 
Episodes ending in combat loss or timeout are counted as failures. 
For each evaluation run, we compute this as 
$\text{Win Rate} = (N_{\mathrm{combat}} + N_{\mathrm{nav}}) / N_{\mathrm{total}}$, 
where $N_{\mathrm{combat}}$ and $N_{\mathrm{nav}}$ denote the number of combat and navigation wins, respectively, and $N_{\mathrm{total}}=32$ for each checkpoint evaluation. 
Since the benchmark intentionally exposes different strategic trade-offs across map variants, we do not aggregate or average win rates across all variants. 
Averaging would obscure the distinct behavioral pressures induced by individual maps and reduce interpretability of agent performance under different tactical conditions.
Our evaluation consists of two complementary components: quantitative and qualitative.

\input{figure/winRate}
\textbf{Quantitative Evaluation:}
We measure training progress using the win-rate metric over time. 
This evaluates whether agents can learn policies that solve either navigation or combat objectives under the benchmark constraints. 
Figure~\ref{fig:win_rate_comparison} compares win-rate trajectories for 3 agents across the full suite.
Across the \textbf{Base} variants, no agent achieves a stable or consistently increasing win rate. 
Performance remains low and fluctuates throughout training, with win rates generally staying below $60\%$. 
This suggests that the equidistant objective layout does not provide an easily exploitable bias toward either navigation or combat, making decisive objective completion difficult for all three agents.
Across the \textbf{Combat-proximal} variants, we observe a similar pattern. 
Although enemies are positioned closer to the allied units, agents still fail to achieve consistent improvement, and win rates remain low and unstable. 
This indicates that shorter-horizon combat availability alone does not guarantee that agents learn reliable combat completion; 
successful combat still requires coordinated movement, targeting, and engagement.
The clearest learning progress appears in the \textbf{Navigation-proximal} variants. 
Both PPO-based agents achieve higher win rates than QMIX in these settings, although their trajectories still exhibit fluctuations. 
MAPPO shows the strongest improvement among the three agents, while MaskPPO and QMIX remain more variable but generally perform better than in the Base and Combat-proximal variants. 
One plausible explanation is that the navigation objective becomes achievable over a shorter horizon in these layouts, making beacon-reaching behavior easier to discover than successful combat. 
% However, we do not draw statistical conclusions from these trends because each setting is evaluated over only three random seeds, and the randomized spawn configurations are independently sampled.
For benchmarking purposes, we recommend interpreting a steadily increasing win-rate curve as evidence that an algorithm is learning a general strategy rather than exploiting a single fixed configuration. 
In addition to overall win rate, we also track success distributions throughout training, allowing us to analyze how objective preference evolves over time and whether agents exhibit bias toward particular objectives. 
Due to the large number of generated plots across algorithms, seeds, and benchmark variants, extended quantitative visualizations are deferred to the Appendix~\ref{appendix:addWinRate}.

\input{figure/termOutDist}
\textbf{Qualitative Evaluation:}
Because Two-Bridge contains mutually exclusive objectives, aggregate win rate alone cannot fully characterize learned behavior.
We therefore examine terminal outcomes from the final policy of each training run.
For each algorithm and variant, the final checkpoint from every random seed is evaluated under the same 32-episode deterministic evaluation protocol, and the resulting terminal outcomes are pooled across seeds.
With three seeds, each distribution is estimated from 96 episodes.
This diagnostic view reveals whether policies favor combat, navigation, or fail through combat loss or timeout under specific strategic conditions.
Figure~\ref{fig:terminal_outcome_distribution} shows that final policies differ not only in how often they succeed, but also in \emph{how} they succeed or fail.
Across the \textbf{Base} variants, all agents are dominated by timeout losses, indicating that policies often fail to reach either objective within the episode limit.
QMIX achieves only minimal navigation wins, while MaskPPO and MAPPO produce slightly more navigation successes.
Combat wins are largely absent, suggesting that agents do not select or complete combat in equidistant layouts.
Across the \textbf{Combat-proximal} variants, timeout losses remain dominant.
Some combat outcomes appear, but mostly as combat losses rather than wins.
Thus, even when combat is available over a shorter horizon, agents do not reliably learn successful combat policies.
In the \textbf{Navigation-proximal} variants, all agents produce more navigation wins than in Base or Combat-proximal settings, especially MAPPO and MaskPPO.
However, persistent timeout losses show that navigation remains nontrivial: 
agents must locate the objective, commit to a route, and move through constrained passages.
Overall, terminal outcome distributions expose objective preferences and failure modes hidden by win rate alone.

%% file: figure/winRate.tex
\begin{figure*}[t]
    \centering
    \includegraphics[width=\textwidth]{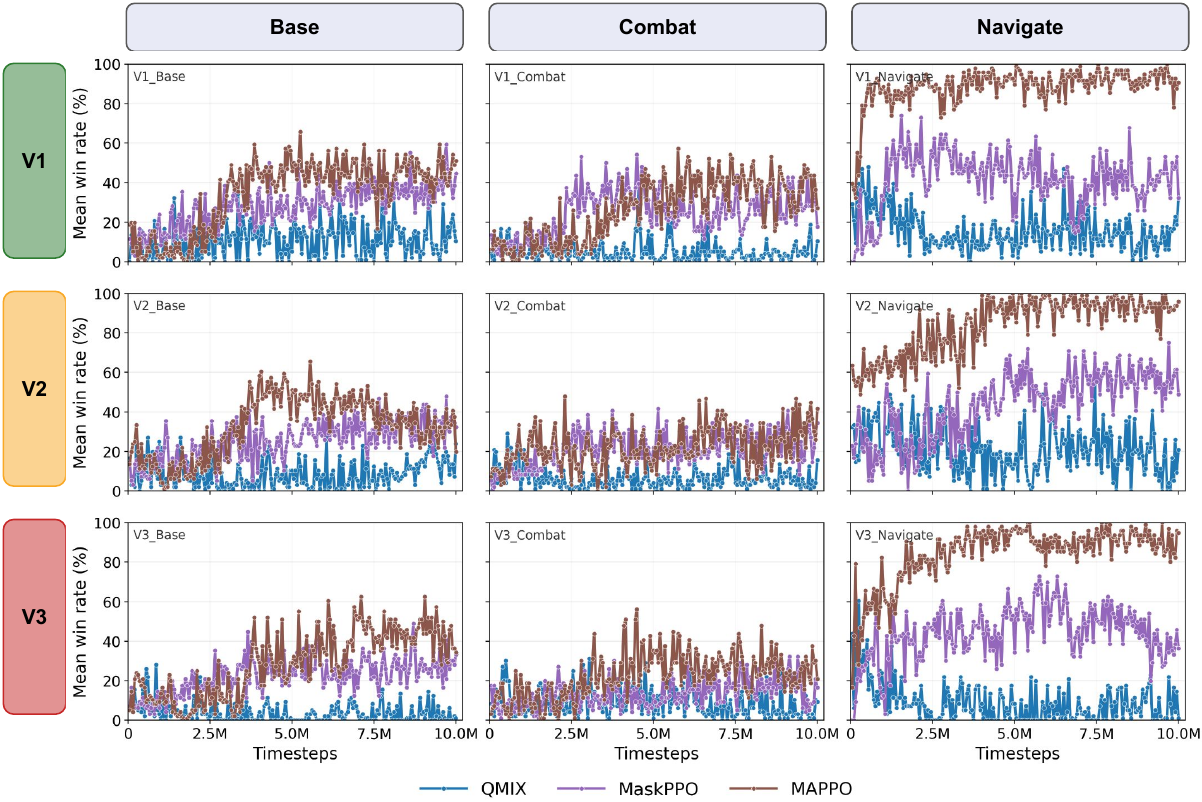}
    \caption{\textbf{Training win-rate trajectories across Two-Bridge variants.}
    Each subplot shows checkpoint-level mean win rate over 10M environment timesteps. 
    Rows indicate unit-count balance variants and columns indicate objective-proximity variants. 
    Performance stays low in Base and Combat-proximal settings, while Navigation-proximal variants show clearer learning progress, particularly for MAPPO and MaskPPO.}
    \label{fig:win_rate_comparison}
\end{figure*}

%% file: figure/termOutDist.tex
\begin{figure*}[t]
    \centering
    \includegraphics[width=\textwidth]{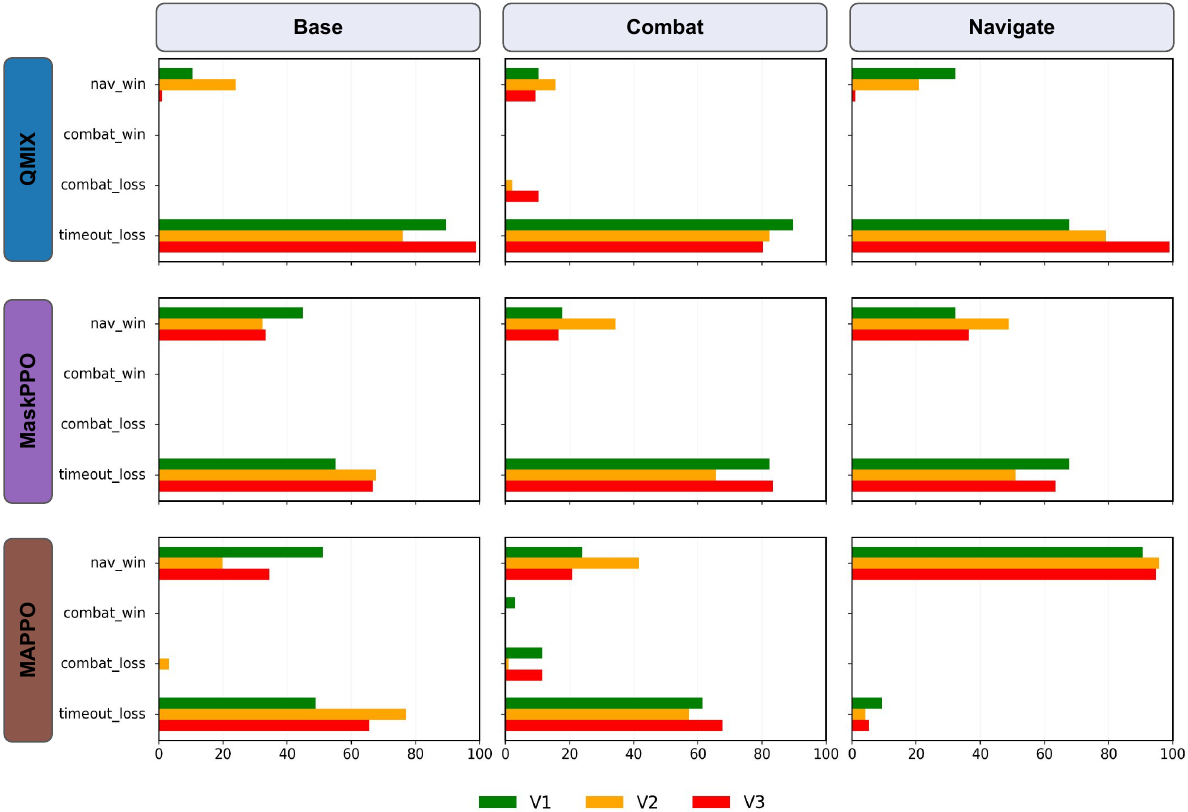}
    \caption{\textbf{Final terminal-outcome distribution across agents and variants.}
    Bars show the percentage of final evaluation episodes ending in navigation win, combat win, combat loss, or timeout loss, pooled across seeds.
    The results reveal that most failures are timeouts, combat wins are rare, and successful policies more often complete navigation, especially in Navigation-proximal settings.}
    \label{fig:terminal_outcome_distribution}
\end{figure*}

%% file: documentBody/4-limAndFut.tex
\section{Discussion, Limitations, and Future Work}

\textbf{Discussion:}
Designed explicitly for modest compute budgets, the Two-Bridge Map Suite nonetheless elicits diverse agent behaviors rather than a single notion of success or failure.
We intended unit-count balance to act as a combat difficulty axis, where friendly-advantage variants make combat easier and enemy-advantage variants make it harder.
However, none of the evaluated algorithms consistently followed this trend: combat wins remained rare even under favorable unit counts.
This suggests that numerical advantage alone is insufficient for learning coordinated engagement, targeting, and execution.
This finding highlights an important property of the benchmark: 
its variants do not merely scale difficulty through surface-level parameters, but expose gaps between environmental affordances and learned policy behavior.
Agents often fail not because success is impossible, but because they do not reliably discover, select, and execute the appropriate objective-specific strategy.
The prevalence of timeout losses further suggests that policies remain indecisive, failing to commit to either navigation or combat within the episode horizon.
At the same time, the stronger performance in Navigation-proximal layouts shows that agents can exploit some structural affordances when the objective becomes easier to discover and complete.
Together, these results suggest that Two-Bridge is useful not only as a performance benchmark, but also as a diagnostic environment for studying strategic failure modes in SC2-based RL.
Many aspects of strategic decision-making remain unresolved, leaving clear headroom for methodological advances beyond incremental tuning.
By emphasizing usability and controlled complexity, Two-Bridge encourages exploration of new representations, learning signals, and coordination strategies under computationally accessible conditions.

\textbf{Limitations:}
The current benchmark is reliably reproducible on Windows and macOS, where SC2 installation through Battle.net is well supported.
Linux support for SC2 and PySC2 remains less stable and inconsistently maintained, which limits deployment on large-scale GPU clusters that are typically Linux-based.
As a result, some large-scale training workflows remain difficult despite the benchmark's clean integration with modern PyTorch-based RL pipelines.
Long-term viability also depends on continued maintenance of external dependencies, including SC2 binaries and PySC2 version compatibility.
These limitations reflect broader infrastructure challenges in SC2-based research rather than issues specific to the Two-Bridge Map Suite.

\textbf{Future Work:}
The Two-Bridge Map Suite enables several natural extensions.
Future work may explore stronger opponents through curriculum learning or self-play.
Reintroducing partial observability would allow investigation of belief-state reasoning.
Additional directions include incorporating unit-type diversity, expanding the objective set, and studying hierarchical or temporally abstract action representations.

%% file: documentBody/5-conclusion.tex
\section{Conclusion}
The Two-Bridge Map Suite provides an open-source, accessible benchmark for studying strategic RL in SC2.
By bridging the gap between existing SC2 benchmarks, it exposes agents to longer horizons, randomized starts, and mutually exclusive navigation and combat objectives.
Our results show that these controlled challenges reveal meaningful behavioral differences and persistent strategic failure modes.
We hope Two-Bridge supports future work on more generalizable, interpretable, and scalable RL methods for complex environments.

%% file: documentBody/Appendix/A-Code.tex
\section{Appendix}\label{appendix:code}

\textbf{Code and Data Availability.}
All map files, Gym-compatible wrappers for the custom SC2 environments, training code, evaluation scripts, agent implementations, and pre-trained model weights are available in the GitHub repository:

\begin{center}
\url{https://github.com/SouravPanda11/Two-Bridge-Map-Suite}
\end{center}

% For ease of access, the repository also includes the dataset used in our experiments. 
% The dataset is additionally archived on Kaggle:

% \begin{center}
% \url{https://kaggle.com/datasets/4295ddd43834889f6574bed4f7aff3af49183ec16889fe2d845e474ac3d6f7d9}
% \end{center}

The repository contains a detailed README with setup and usage instructions. 
Together, the code, maps, datasets, and pre-trained models provide an end-to-end framework for reproducing our results, running new experiments, and extending the Two-Bridge benchmark with minimal setup effort.

\textbf{Hardware Configuration}\label{hardware}
All experiments were conducted on a single workstation equipped with an Intel Core i5-13600K CPU (13th Gen), 32 GB of RAM, and an NVIDIA RTX A4500 GPU with 20 GB of VRAM.

\section{Training Details}
\label{appendix:training}

\subsection{Reduced Two-Bridge observation}
The reduced Two-Bridge observation consists of a spatial minimap component and a compact non-spatial unit-status vector. 
The spatial component is obtained from the PySC2 feature minimap. 
Although the feature minimap is requested at $64 \times 64$, only the playable Two-Bridge corridor is exposed to the learning algorithms. 
Specifically, the minimap is cropped using the window
\(
(y_0,y_1,x_0,x_1)=(28,60,0,32),
\)
which produces a $32 \times 32$ spatial input. 
In the standard reduced setting, the minimap has two channels,
\(
M_t \in \mathbb{N}^{2 \times 32 \times 32}.
\)
The first channel is the PySC2 pathability layer, which identifies walkable and blocked terrain. 
The second channel is the player-relative layer, which gives a coarse categorical map of empty, allied, neutral, and enemy regions.

The vector component records unit status without exposing exact unit coordinates. 
For each friendly and enemy unit slot, it stores health and an alive indicator. 
It also includes elapsed game time and the number of remaining enemies. 
With $N_F=5$ friendly units and $N_E$ enemies, the actor-facing vector is
\(
o^{\mathrm{vec}}_t =
[
h^F_1,\ell^F_1,\ldots,h^F_{N_F},\ell^F_{N_F},
h^E_1,\ell^E_1,\ldots,h^E_{N_E},\ell^E_{N_E},
\tau_t,n^E_t
],
\)
where $h$ is health, $\ell \in \{0,1\}$ is the alive flag, $\tau_t$ is elapsed game time, and $n^E_t$ is the number of alive enemies. 
Therefore,
\(
\dim(o^{\mathrm{vec}}_t)=2N_F+2N_E+2.
\)
For V1, V2, and V3, the enemy counts are $N_E=3,5,8$, giving actor-vector dimensions $18$, $22$, and $28$, respectively.

\textbf{Single-agent MaskPPO NS reduced:}
MaskPPO NS reduced uses a centralized single-agent interface. 
The policy receives the reduced minimap and the compact vector observation as a single dictionary observation. 
In the V2 combat setting, this corresponds to
\(
M_t \in \mathbb{N}^{2 \times 32 \times 32}, \qquad
o^{\mathrm{vec}}_t \in \mathbb{R}^{22}.
\)
The policy outputs a joint action containing one discrete action for each of the five friendly units.

\textbf{Multi-agent MAPPO and QMIX reduced:}
MAPPO reduced and QMIX reduced use a multi-agent interface with one agent per friendly unit. 
Each of the five agents receives the same compact actor-facing vector observation, while the reduced minimap is provided as a shared spatial input and encoded separately by the neural network. 
Thus the per-agent observation matrix has shape
\(
O_t \in \mathbb{R}^{N_F \times (2N_F+2N_E+2)}.
\)
A one-hot agent identifier is appended internally by the MAPPO and QMIX policies so that the shared policy can distinguish friendly unit slots.

During centralized training, MAPPO and QMIX also use a global training state derived from raw-unit information. 
This centralized state contains friendly and enemy coordinates, health values, alive flags, beacon position, minimum friendly-to-beacon distance, elapsed time, and remaining enemy count. 
This state is used for centralized value estimation or mixing, but it is distinct from the actor-facing observation described above.

\subsection{Reduced Two-Bridge action space}
All reduced agents use the same high-level unit-command action interface. 
At each environment step, the controller selects one discrete action for each friendly unit slot. 
There are always $N_F=5$ friendly unit slots. 
The number of enemy slots, $N_E$, depends on the map variant: 
$N_E=3$ for V1, $N_E=5$ for V2, and $N_E=8$ for V3.

For each friendly unit $i$, the per-unit action space is
\[
\mathcal{A}_i =
\{\text{no-op}\}
\cup
\{\text{move in one of eight directions}\}
\cup
\{\text{attack enemy slot }1,\ldots,N_E\}.
\]
Therefore, the number of discrete actions per friendly unit is
\(
|\mathcal{A}_i| = 1 + 8 + N_E.
\)
This gives $12$ actions per unit in V1, $14$ actions per unit in V2, and $17$ actions per unit in V3.

The action IDs are ordered as follows:
\(
a_i = 0
\)
corresponds to no-op. 
Actions
\(
a_i \in \{1,\ldots,8\}
\)
are relative movement commands. 
Each movement action moves the selected friendly unit by a fixed step of two raw-coordinate units:
\[
\begin{aligned}
1 &: (0,-2), &
2 &: (0,+2), &
3 &: (-2,0), &
4 &: (+2,0), \\
5 &: (+2,-2), &
6 &: (-2,-2), &
7 &: (+2,+2), &
8 &: (-2,+2).
\end{aligned}
\]
The movement target is clipped to the valid raw map bounds before being sent to PySC2. 
Movement legality is not masked using pathability; 
all eight movement actions are available for alive friendly units. 
Terrain constraints are instead handled by the SC2 engine, and the policy must learn navigability from the minimap observation.

Attack actions begin after the movement actions. Let
\(
a_{\mathrm{attack}} = 1+8 = 9.
\)
Then action
\(
a_i = a_{\mathrm{attack}} + j
\)
commands friendly unit $i$ to attack enemy slot $j$, where $j \in \{0,\ldots,N_E-1\}$. 
Enemy slots are maintained using stable PySC2 unit tags, so an attack action refers to an enemy slot rather than to a raw coordinate. 
If the corresponding enemy unit is dead, missing, or outside attack range, the action is unavailable and is treated as invalid by the environment.

Each environment step submits the selected per-unit commands simultaneously. Thus a joint action is
\(
\mathbf{a}_t =
(a^1_t,\ldots,a^{N_F}_t),
\)
where each component controls one friendly unit slot. 
If a friendly unit is dead or missing, its requested action is ignored. 
If no valid unit command is produced by the joint action, the environment sends a global PySC2 no-op command.

\textbf{Single-agent MaskPPO NS reduced action interface:}
MaskPPO NS reduced uses a centralized single-agent controller that outputs one joint action for all five friendly units. 
The Gym action space is a multi-discrete space,
\[
\mathcal{A}^{\mathrm{MaskPPO}}
=
\mathrm{MultiDiscrete}
\left(
[1+8+N_E,\ldots,1+8+N_E]
\right),
\]
with one branch per friendly unit. 
In the V2 combat setting used by MaskPPO NS reduced, $N_E=5$, so the action space is
\(
\mathrm{MultiDiscrete}([14,14,14,14,14]).
\)
Although the policy is trained as a single centralized policy, the semantics remain per-unit: 
each of the five discrete branches selects the command for one friendly unit slot.

\textbf{Multi-agent MAPPO and QMIX reduced action interface:}
MAPPO reduced and QMIX reduced use a multi-agent action interface with one learning agent per friendly unit. 
Each agent has the same discrete action set,
\(
\mathcal{A}_i = \{0,\ldots,8+N_E\}.
\)
The environment therefore exposes
\(
N_F=5
\)
agents, each with
\(
n_{\mathrm{actions}} = 1+8+N_E
\)
available action indices. 
For V1, V2, and V3 this gives $12$, $14$, and $17$ actions per agent, respectively. 
The full environment transition still depends on the joint action across all five agents,
\(
\mathbf{a}_t =
(a^1_t,a^2_t,a^3_t,a^4_t,a^5_t),
\)
but MAPPO and QMIX factor the decision across the five friendly unit controllers.

\textbf{Action availability masks:}
The implementation also provides action-availability masks. 
These masks are part of the action interface rather than the semantic observation features. 
For each friendly unit, no-op is always available. 
If the friendly unit is alive, all eight movement actions are available. 
Attack actions are available only when the corresponding enemy slot is alive and within attack range. 
The attack range check uses the internal raw-unit positions and a Euclidean distance threshold of $6.0$ raw-coordinate units.

For MAPPO and QMIX reduced, the availability mask has shape
\(
A^{\mathrm{avail}}_t
\in
\{0,1\}^{N_F \times (1+8+N_E)}.
\)
MAPPO uses this mask to set unavailable action logits to a large negative value before sampling from the categorical policy. 
QMIX uses the same mask during action selection and target computation so that unavailable actions are not selected by the Q-network.
For MaskPPO NS reduced, the environment initially produces the same per-unit mask shape. 
The training wrapper flattens it for compatibility with MaskablePPO:
\(
A^{\mathrm{mask}}_t
\in
\{0,1\}^{N_F(1+8+N_E)}.
\)
In the V2 combat setting, this is a length-$70$ mask because $N_F=5$ and $1+8+N_E=14$.

\subsection{Reduced Two-Bridge reward}
All reduced agents are trained with a shared team reward. 
The reward is computed after each joint action is executed in the SC2 environment and is returned as a single scalar:
\(
r_t = r^{\mathrm{nav}}_t + r^{\mathrm{combat}}_t + r^{\mathrm{term}}_t .
\)
The same scalar reward is used for all controlled friendly units. 
Thus, the reward is cooperative: it evaluates team progress rather than assigning separate local rewards to individual units.

The reward has three components. 
The navigation component encourages the friendly units to approach the beacon objective. 
Let \(p^F_{i,t}\) be the position of friendly unit \(i\), let \(b_t\) be the beacon position, and let \(\mathcal{F}_t\) be the set of alive friendly units. 
For each alive friendly unit, the environment computes its Euclidean distance to the beacon,
\(
d^B_{i,t} = \|p^F_{i,t} - b_t\|_2 .
\)
The navigation shaping reward is the mean reduction in beacon distance among alive friendly units:
\[
r^{\mathrm{nav}}_t
=
\frac{1}{|\mathcal{F}_t|}
\sum_{i \in \mathcal{F}_t}
\left(
d^B_{i,t-1} - d^B_{i,t}
\right).
\]
This term is positive when friendly units move closer to the beacon and negative when they move farther away.

The combat component encourages the friendly units to approach and damage enemies. 
When at least one enemy is alive, the environment computes the centroid of the alive enemy units:
\(
c^E_t
=
\frac{1}{|\mathcal{E}_t|}
\sum_{j \in \mathcal{E}_t}
p^E_{j,t},
\)
where \(\mathcal{E}_t\) is the set of alive enemies. 
The combat-distance shaping term is
\[
r^{\mathrm{centroid}}_t
=
\frac{1}{|\mathcal{F}_t|}
\sum_{i \in \mathcal{F}_t}
\left(
\|p^F_{i,t-1} - c^E_{t-1}\|_2
-
\|p^F_{i,t} - c^E_t\|_2
\right).
\]
This term rewards the friendly group for moving closer to the current enemy formation.

The health-based combat reward rewards enemy damage and penalizes friendly damage. 
Let \(H^E_t\) and \(H^F_t\) denote total enemy and friendly health at timestep \(t\). 
The health shaping scale is
\(
\lambda_{\mathrm{hp}} = 0.05 .
\)
Conceptually, the health term is
\(
r^{\mathrm{hp}}_t
=
\lambda_{\mathrm{hp}}
\left(H^E_{t-1} - H^E_t\right)
-
\lambda_{\mathrm{hp}}
\left(H^F_{t-1} - H^F_t\right).
\)
Thus, enemy health reduction gives positive reward, while friendly health reduction gives negative reward.

Enemy kills provide an additional bonus. With
\(
\lambda_{\mathrm{kill}} = 1.0,
\)
the kill reward is
\(
r^{\mathrm{kill}}_t
=
\lambda_{\mathrm{kill}}
\cdot
K^E_t ,
\)
where \(K^E_t\) is the number of enemies that were alive at \(t-1\) and dead at \(t\).

The combat reward is therefore
\(
r^{\mathrm{combat}}_t
=
r^{\mathrm{centroid}}_t
+
r^{\mathrm{hp}}_t
+
r^{\mathrm{kill}}_t .
\)
For the MaskPPO NS reduced implementation, friendly unit deaths also receive an explicit loss penalty,
\(
-\lambda_{\mathrm{kill}} K^F_t,
\)
where \(K^F_t\) is the number of friendly units that died during the transition. 
MAPPO reduced and QMIX reduced do not include this separate friendly-death count term; 
friendly losses are penalized through the friendly health-loss component and terminal loss penalty.

The terminal reward is added only when the episode ends:
\[
r^{\mathrm{term}}_t =
\begin{cases}
+25, & \text{navigation win},\\
+10, & \text{combat win},\\
-10, & \text{combat loss},\\
-15, & \text{timeout loss},\\
0, & \text{tie}.
\end{cases}
\]
A navigation win occurs when the minimum distance between any alive friendly unit and the beacon is less than the beacon radius,
\(
\min_{i \in \mathcal{F}_t} d^B_{i,t} < 5.0 .
\)
A combat win occurs when all enemies are dead. 
A combat loss occurs when all friendly units are dead. 
A timeout loss occurs when the episode reaches the environment step limit without completing either objective. 
The default episode limit corresponds to five minutes of SC2 game time, or \(600\) environment steps with step multiplier \(8\).

\textbf{MaskPPO NS reduced reward:}
MaskPPO NS reduced uses the shared team reward described above in a centralized single-agent interface. 
The policy outputs a joint action for all five friendly units, and the environment returns one scalar reward for the resulting team transition. 
The MaskPPO NS reduced reward includes navigation shaping, enemy-centroid combat shaping, enemy damage reward, friendly damage penalty, enemy kill bonus, and an explicit friendly death penalty. The scalar reward can be written as
\[
r_t =
r^{\mathrm{nav}}_t
+
r^{\mathrm{centroid}}_t
+
0.05(H^E_{t-1}-H^E_t)
-
0.05(H^F_{t-1}-H^F_t)
+
K^E_t
-
K^F_t
+
r^{\mathrm{term}}_t .
\]
This reward is returned once per environment step and is assigned to the centralized MaskPPO rollout transition.

\textbf{MAPPO reduced reward:}
MAPPO reduced uses the same cooperative team reward for all five agents. 
Although the action interface is multi-agent, with one agent per friendly unit, the environment does not return separate individual rewards. 
Instead, each timestep produces a single scalar team reward that is shared across the agents for policy-gradient updates.
MAPPO reduced follows the reduced QMIX-style reward implementation. 
The reward includes navigation progress, enemy-centroid approach, enemy health reduction, friendly health loss penalty, enemy kill bonus, and terminal reward:
\(
r_t =
r^{\mathrm{nav}}_t
+
r^{\mathrm{centroid}}_t
+
0.05\Delta H^E_t
-
0.05\Delta H^F_t
+
K^E_t
+
r^{\mathrm{term}}_t .
\)
Here,
\(
\Delta H^E_t = \sum_j \max(h^E_{j,t-1}-h^E_{j,t},0),
\)
and
\(
\Delta H^F_t = \sum_i \max(h^F_{i,t-1}-h^F_{i,t},0).
\)
The centralized critic uses the global state and minimap embedding to estimate the value of this shared team reward.

\textbf{QMIX reduced reward}
QMIX reduced uses the same reward as MAPPO reduced. 
Each friendly unit is treated as an individual agent for action selection, but the learning objective is fully cooperative. 
The environment returns one scalar team reward, and QMIX uses this reward as the target signal for the joint action-value function produced by the mixing network.
As in MAPPO reduced, the reward is
\(
r_t =
r^{\mathrm{nav}}_t
+
r^{\mathrm{centroid}}_t
+
0.05\Delta H^E_t
-
0.05\Delta H^F_t
+
K^E_t
+
r^{\mathrm{term}}_t .
\)
The individual agent Q-values are therefore not trained against separate per-unit rewards. 
Instead, they are mixed into a centralized joint value that is optimized against the shared team return.

\textbf{Implementation note:}
On the first reward computation after reset, the previous-distance and previous-health buffers are initialized and the shaping reward is set to zero. 
This avoids assigning an artificial reward before a valid transition history exists.

\subsection{Reduced-agent training hyperparameters}
All reduced-agent experiments were trained for 10M environment steps per seed, using 3 random seeds and 3 parallel SC2 environment workers unless otherwise stated. 
Model checkpoints were saved every 50K environment steps. 
Training was performed without SC2 visualization or realtime execution. 
TensorBoard logging was enabled for the reduced MAPPO and QMIX runs and for the MaskPPO reward-component logger.

\begin{table}[h]
\centering
\caption{Shared training settings for reduced agents.}
\begin{tabular}{ll}
\toprule
Hyperparameter & Value \\
\midrule
Total environment steps per seed & 10M \\
Number of seeds & 3 \\
Parallel environments & 3 \\
Checkpoint interval & 50K steps \\
SC2 visualization & Disabled \\
SC2 realtime mode & Disabled \\
Device & CUDA if available, otherwise CPU \\
\bottomrule
\end{tabular}
\end{table}

\textbf{MaskPPO NS reduced hyperparameters:}
MaskPPO NS reduced was trained with the \texttt{MaskablePPO} implementation from \texttt{sb3-contrib}, using the \texttt{MultiInputPolicy}. 
The policy receives the reduced minimap, compact vector observation, and action mask through the multi-input environment interface. 
The action mask is also exposed through the MaskablePPO masking API to prevent invalid actions during sampling.

\begin{table*}[t]
\centering
\caption{Training hyperparameters for reduced-agent experiments.}
\scriptsize
\setlength{\tabcolsep}{3pt}
\renewcommand{\arraystretch}{1.08}
\begin{tabular}{p{0.23\textwidth}p{0.24\textwidth}p{0.24\textwidth}p{0.24\textwidth}}
\toprule
Hyperparameter & MaskPPO NS reduced & MAPPO reduced & QMIX reduced \\
\midrule
Algorithm & MaskablePPO & MAPPO & QMIX \\
Training interface & Centralized joint-action controller & Multi-agent actor with centralized critic & Multi-agent value decomposition \\
Policy/network type & \texttt{MultiInputPolicy} & Shared actor and centralized critic & Shared per-agent Q-network and QMIX mixer \\
Environment steps per seed & 10M & 10M & 10M \\
Number of seeds & 3 & 3 & 3 \\
Parallel SC2 workers & 3 & 3 & 3 \\
Checkpoint interval & 50K steps & 50K steps & 50K steps \\
SC2 visualization / realtime & Disabled / disabled & Disabled / disabled & Disabled / disabled \\
Reduced minimap channels & Pathability + player-relative & Pathability + player-relative & Pathability + player-relative \\
Reduced minimap shape & $2 \times 32 \times 32$ & $2 \times 32 \times 32$ & $2 \times 32 \times 32$ \\
Rollout length & $512$ steps per worker & $512$ steps & -- \\
Rollout size per update & $512 \times 3 = 1536$ transitions & $512 \times 3 = 1536$ transitions & -- \\
Minibatch / batch size & $512$ transitions & $256$ transitions & $16$ episodes \\
Update epochs & $4$ PPO epochs & $4$ PPO epochs & -- \\
Replay buffer size & -- & -- & $5000$ episodes \\
Learning starts after & -- & -- & $32$ episodes \\
Training updates per episode & -- & -- & $1$ \\
Learning rate & $3 \times 10^{-4}$ & $3 \times 10^{-4}$ & $5 \times 10^{-4}$ \\
Optimizer & Adam & Adam & Adam \\
Discount factor $\gamma$ & $0.99$ & $0.99$ & $0.99$ \\
GAE parameter $\lambda$ & $0.95$ & $0.95$ & -- \\
PPO clip range & $0.2$ & $0.2$ & -- \\
Entropy coefficient & $0.0$ & $0.001$ & -- \\
Value-loss coefficient & $0.5$ & $0.5$ & -- \\
Maximum gradient norm & $0.5$ & $10.0$ & $10.0$ \\
Advantage normalization & Enabled & Enabled & -- \\
Reward standardization & Not used & Enabled & Enabled \\
Target network update interval & -- & -- & $200$ training updates \\
Double Q-learning & -- & -- & Enabled \\
Exploration schedule & PPO sampling & PPO sampling & $\epsilon: 1.0 \rightarrow 0.05$ over $50{,}000$ steps \\
Hidden dimension & $[64,64]$ policy/value MLPs & $128$ actor and critic hidden dimension & $64$ shared Q-network hidden dimension \\
Activation function & Tanh & ReLU & ReLU \\
Orthogonal initialization & Enabled & -- & -- \\
Feature/minimap encoder & SB3 \texttt{CombinedExtractor} & CNN minimap encoder, $64$-dim. output & CNN minimap encoder, $64$-dim. output \\
Use minimap in actor/Q-network & Enabled & Enabled & Enabled \\
Use minimap in critic/mixer & Enabled & Enabled & Enabled \\
Append one-hot agent ID & Not applicable & Enabled & Enabled \\
Recurrent network & Not used & Not used & Disabled \\
Mixer embedding dimension & -- & -- & $32$ \\
Hypernetwork layers & -- & -- & $2$ \\
Hypernetwork hidden dimension & -- & -- & $64$ \\
Evaluation during training & Disabled & Disabled & Disabled \\
TensorBoard logging & Reward components & Enabled & Enabled \\
Device & CUDA if available, otherwise CPU & CUDA if available, otherwise CPU & CUDA if available, otherwise CPU \\
\bottomrule
\end{tabular}
\end{table*}

For MAPPO reduced and QMIX reduced, the reduced minimap encoder is
\(
\mathrm{Conv}(2,16,k=5,s=2,p=2)
\rightarrow
\mathrm{ReLU}
\rightarrow
\mathrm{Conv}(16,32,k=3,s=2,p=1)
\rightarrow
\mathrm{ReLU}
\rightarrow
\mathrm{Conv}(32,32,k=3,s=2,p=1)
\rightarrow
\mathrm{ReLU}
\rightarrow
\mathrm{Linear}(\cdot,64).
\)

\paragraph{MAPPO reduced hyperparameters.}
MAPPO reduced was trained with a shared actor and centralized critic. 
Each friendly unit is treated as an agent, and a one-hot agent identifier is appended internally to distinguish unit slots. 
The reduced minimap is encoded by a convolutional encoder and appended to both the actor and critic inputs in the reported reduced MAPPO runs.

\textbf{QMIX reduced hyperparameters:}
QMIX reduced was trained with a shared per-agent Q-network and a centralized QMIX mixing network. 
Each friendly unit corresponds to one agent. 
The shared Q-network receives the per-agent vector observation, a one-hot agent identifier, and a shared minimap embedding. 
The mixer receives the centralized state concatenated with the same minimap embedding.

%% file: documentBody/Appendix/B-mainResults.tex
\section{Additional Results}
\label{appendix:addWinRate}
\input{figure/Appendix/MainResults/qmixStacked}
\input{figure/Appendix/MainResults/maskppoStacked}
\input{figure/Appendix/MainResults/mappoStacked}

%% file: figure/Appendix/MainResults/qmixStacked.tex
\begin{figure}[h]
    \centering
    \includegraphics[width=\textwidth]{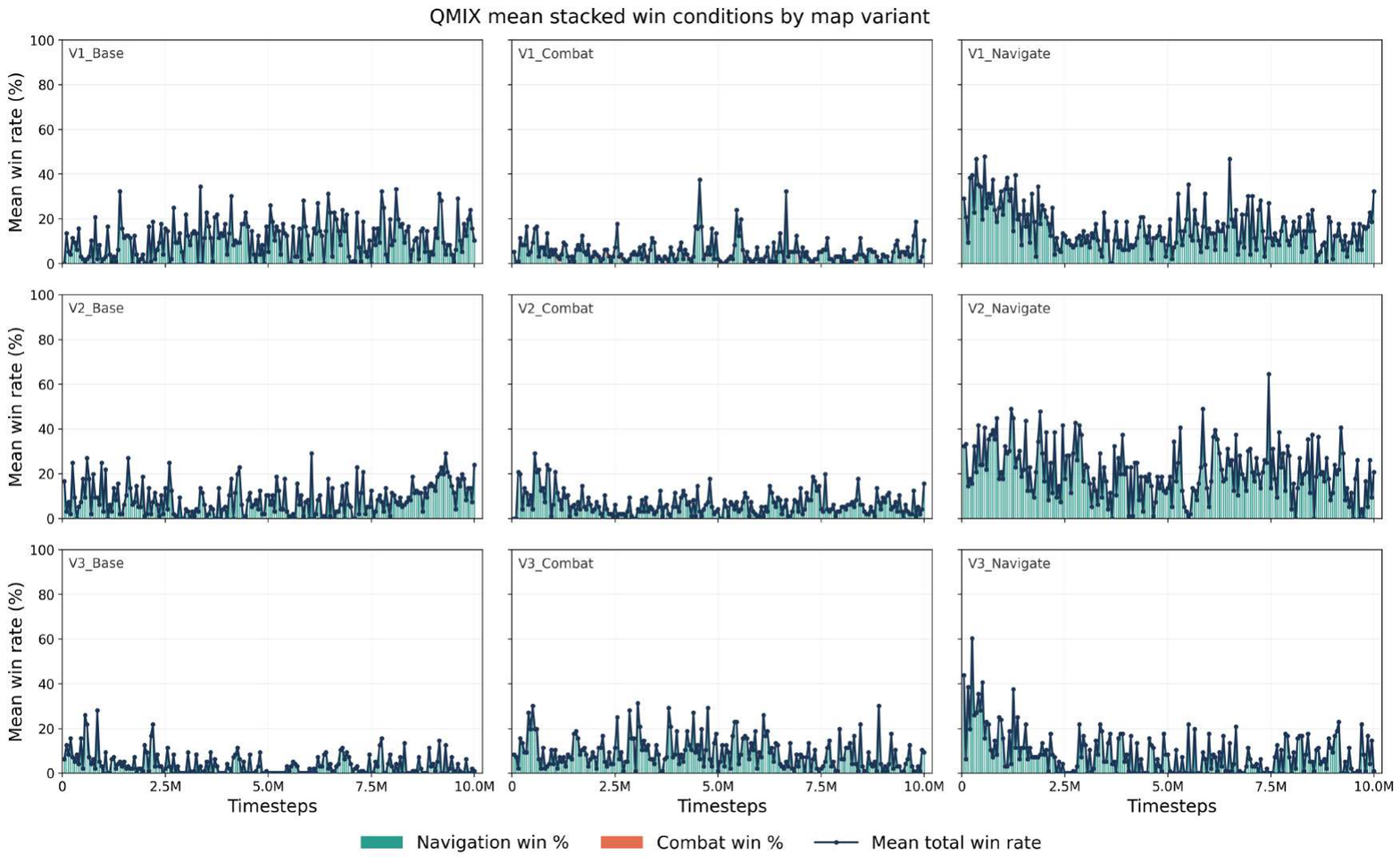}
    \caption{QMIX mean stacked win conditions}
    \label{fig:qmix_stacked}
\end{figure}

%% file: figure/Appendix/MainResults/maskppoStacked.tex
\begin{figure}[h]
    \centering
    \includegraphics[width=\textwidth]{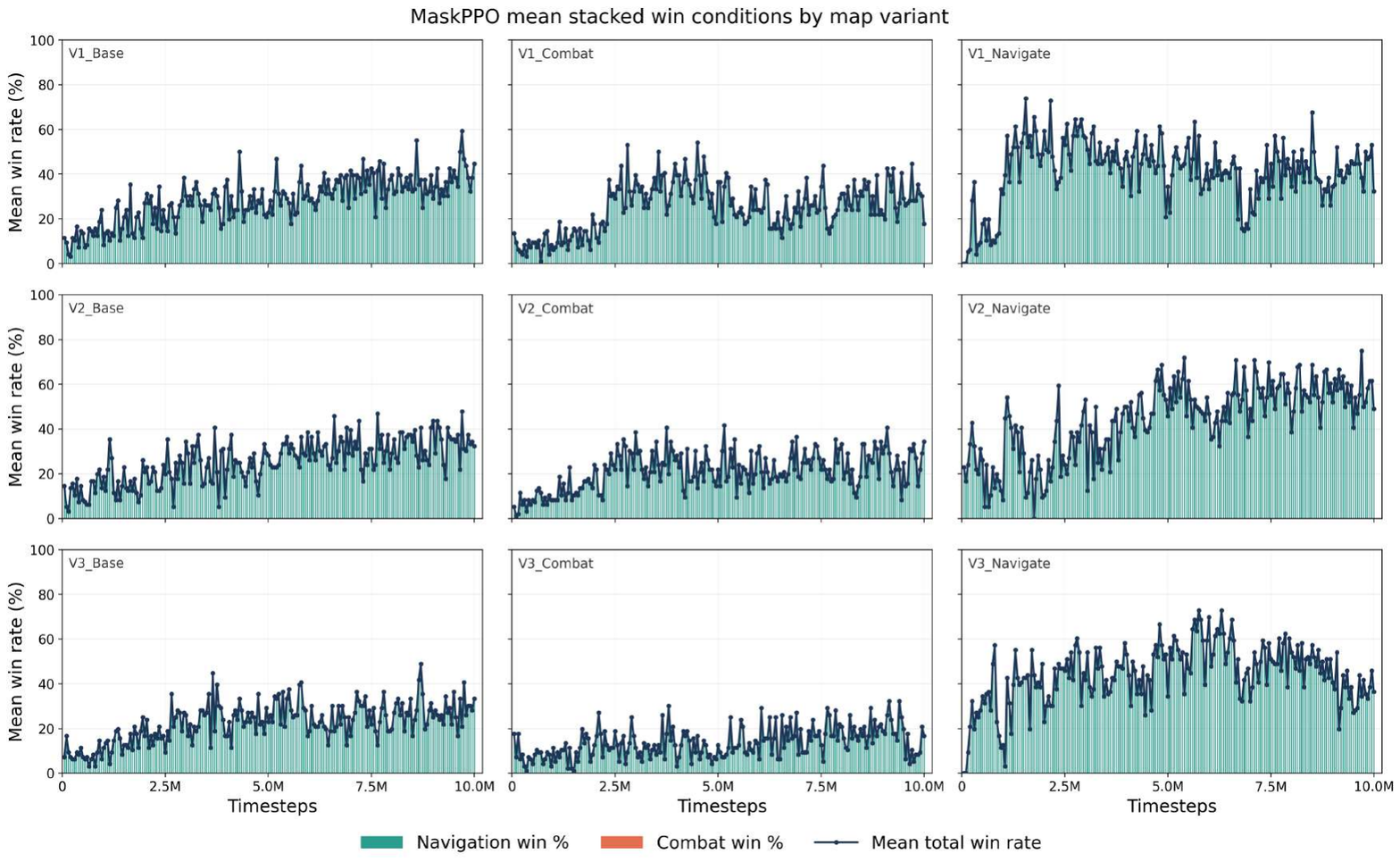}
    \caption{MaskPPO mean stacked win conditions}
    \label{fig:maskPPO_stacked}
\end{figure}

%% file: figure/Appendix/MainResults/mappoStacked.tex
\begin{figure}[h]
    \centering
    \includegraphics[width=\textwidth]{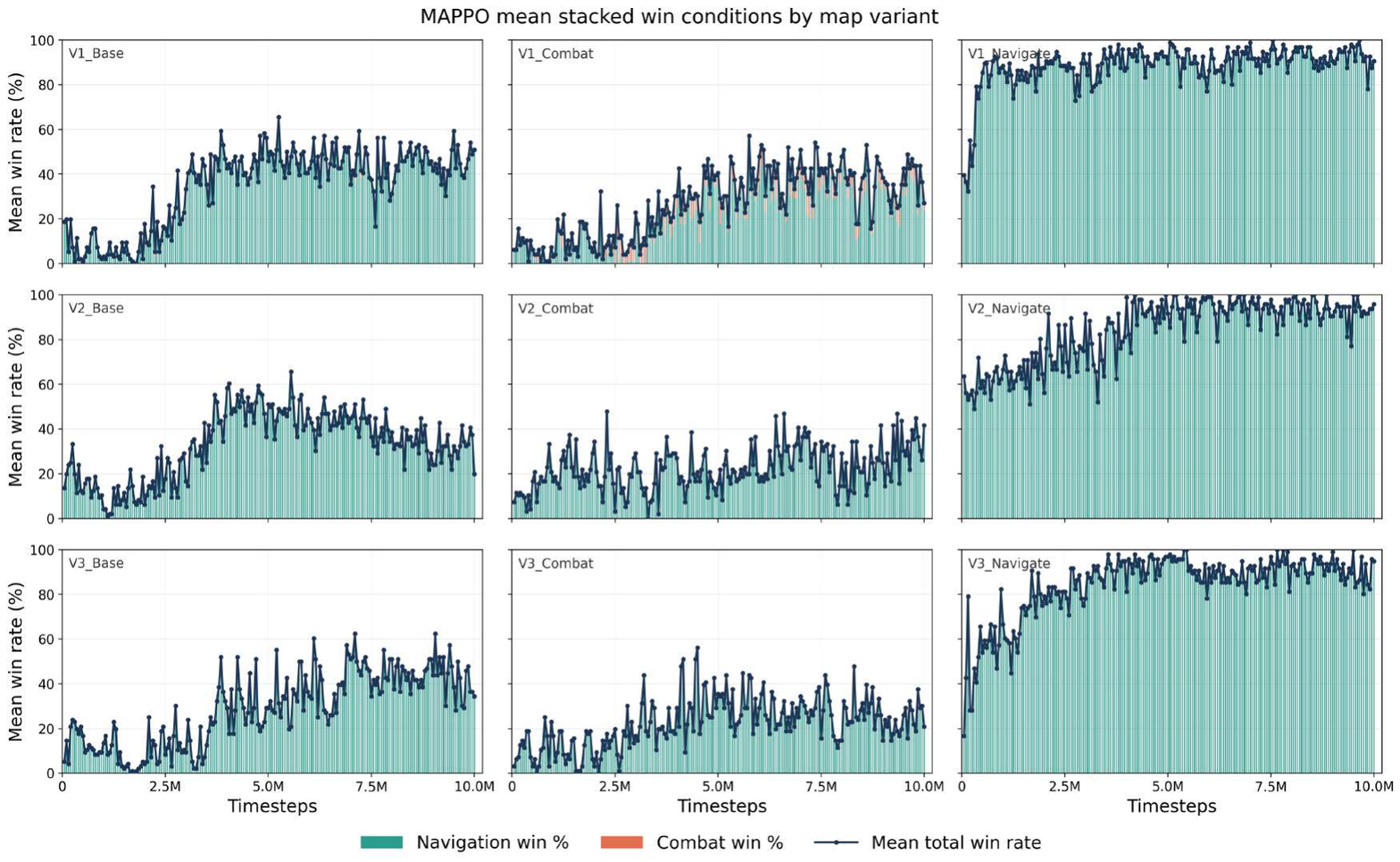}
    \caption{MAPPO mean stacked win conditions}
    \label{fig:mappo_stacked}
\end{figure}

%% file: documentBody/Appendix/C.1-pilot1.tex
\section{Pilot 1: Preliminary Training Strategy}
\label{appendix:pilot1}

Before scaling to the full Two-Bridge benchmark suite, we conducted an initial pilot study on the \texttt{V2\_Base} variant, a balanced unit-count configuration used as a representative mid-difficulty setting. 
The goal was not to establish benchmark-level performance, but to identify early limitations in the observation space, action interface, and reward design under a preliminary environment setup.

\textbf{Observation Space:}
We evaluate two PySC2-based observation modalities: \textbf{Non-Spatial Features (NSF)} and \textbf{Spatial Features (SF)}.
In NSF, the agent receives a 55-dimensional vector containing unit metadata, including positions, health, weapon cooldowns, beacon coordinates, distance to the beacon, elapsed time, and remaining enemy count.
This representation is lightweight but lacks explicit spatial structure.
In SF, this vector is augmented with PySC2 spatial inputs: a 17-channel \texttt{feature\_screen} and a 7-channel \texttt{feature\_minimap}, each at $64 \times 64$ resolution, providing additional terrain and unit-distribution information.

\textbf{Action Space:}
Both observation settings use the same discrete unit-level action space.
Each friendly Marine selects one of 14 commands: no-op, eight directional movements, or one of five enemy-targeted attack actions.

\textbf{Reward Function:}
The pilot reward combines navigation, combat, and terminal feedback:
\( r_t = r^{\text{nav}}_t + r^{\text{combat}}_t + r^{\text{terminal}} \).
Navigation progress is measured by the change in Euclidean distance between a designated leading unit and the beacon:
\( r^{\text{nav}}_t = d_{t-1} - d_t \).
The leading unit is the first friendly Marine under a fixed unit-tag ordering, and the reward is shared across all controlled units.
Combat progress is defined as a kill--loss delta:
\( r^{\text{combat}}_t = (E_{t-1} - E_t) - (F_{t-1} - F_t) \),
where \(E_t\) and \(F_t\) are the numbers of living enemy and friendly units at timestep \(t\).
Terminal rewards are \(+10\) for beacon capture or enemy elimination, \(-10\) for friendly elimination or timeout, and \(0\) for ties.

\textbf{Training Details:}
We train PPO~\cite{schulman2017proximal} and A2C~\cite{mnih2016asynchronous} using Stable-Baselines3~\cite{stable-baselines3} with default hyperparameters for 2M environment timesteps.
NSF experiments use \texttt{MlpPolicy}, while SF experiments use \texttt{MultiInputPolicy}.

\input{figure/Appendix/C_pilot1Results}
\subsection{Pilot 1: Results and Analysis}
We analyze terminal outcomes and qualitative policy behavior across PPO and A2C under NSF and SF observations. 
Figure~\ref{fig:exp1_results} reports terminal outcomes and TensorBoard logs for all four settings. 
Across runs, the dominant failure mode is single-unit control: policies primarily command one active unit at a time and switch only after that unit is lost.

\textbf{PPO-NSF} learns a degenerate fixed-direction policy, repeatedly sending one unit toward the upper-right regardless of objective placement. 
Wins occur only when the beacon is incidentally encountered, while enemy encounters usually lead to combat loss and repeated unit sacrifice. 
\textbf{PPO-SF} shows slightly richer movement, using a short rightward sweep followed by an upward scan, but still behaves mainly as a beacon-search policy. 
Enemy encounters often result in leading-unit loss, after which remaining units are routed toward the bottom-right and frequently idle until timeout.

\textbf{A2C-NSF} fails to learn a stable conditional strategy and often spends excessive time traversing the central terrain constraint. 
After crossing, it shows no consistent preference between navigation and combat, producing varied terminal outcomes and frequent timeouts. 
\textbf{A2C-SF} collapses to deterministic parking behavior: the agent crosses the nearest bridge, moves toward the right boundary at a fixed vertical offset, and remains stationary. 
Beacon encounters along this path produce incidental navigation wins, while enemy encounters usually trigger repeated unit loss and timeout.

%% file: figure/Appendix/C_pilot1Results.tex
\begin{figure}[h]
    \centering
    \includegraphics[width=\textwidth]{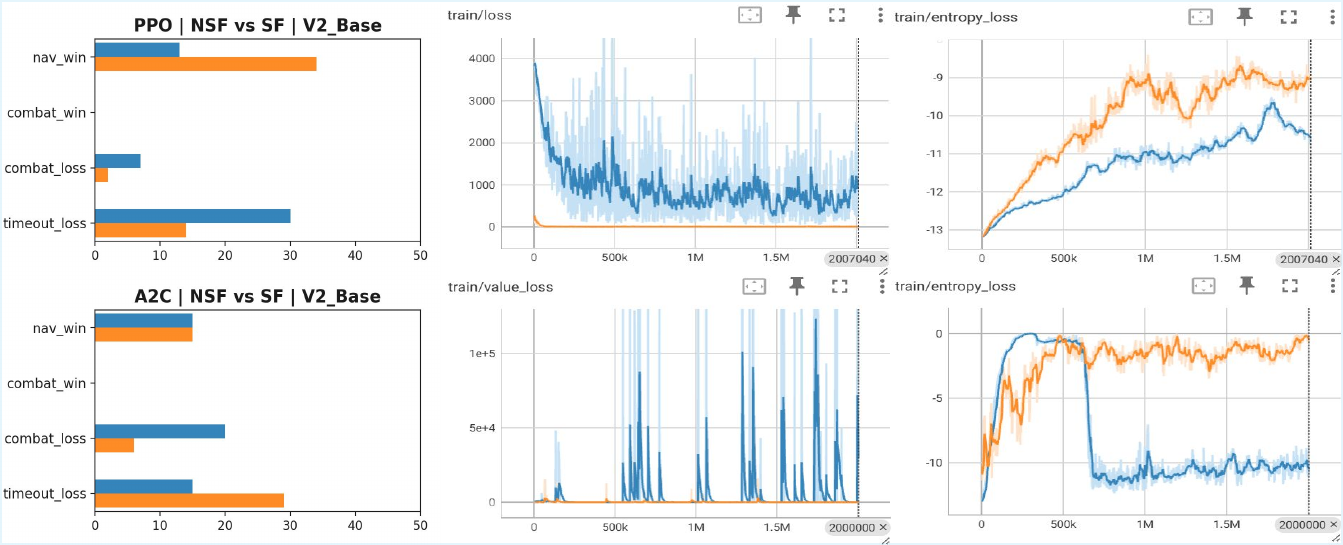}
    \caption{
    Experiment 1: Terminal outcome distribution and training dynamics for PPO and A2C under NSF and SF observations on V2\_Base.
    Training curves are shown with smoothing factor 0.8.
    }
    \label{fig:exp1_results}
\end{figure}

%% file: documentBody/Appendix/C.2-pilot23.tex
\section{Pilot 2: Action Masking}
\label{appendix:pilot2}

Pilot 1 was used only to identify preliminary design limitations. 
Pilot 2 defines a self-contained training setup for evaluation across the benchmark suite, using a fixed observation representation, structured action interface, action masking, and revised reward design.

\textbf{Observation Space.}
We use a combined vector--spatial observation from \texttt{PySC2}. 
The vector component is a 55-dimensional representation encoding unit metadata, beacon information, elapsed time, and remaining enemy count. 
The spatial component includes a 17-channel \texttt{feature\_screen} and a 7-channel \texttt{feature\_minimap}, each at $64 \times 64$ resolution, providing terrain and unit-distribution context. 
This combined representation is fixed for all Pilot 2 experiments.

\textbf{Action Space.}
We use a centralized, unit-level discrete action space. 
At each timestep, the agent selects a high-level verb (\emph{no-op}, \emph{move}, or \emph{attack}), a binary mask over friendly units, a movement direction, and, for attacks, an enemy target index. 
Each Marine can execute one of $9 + N_E$ commands: no-op, eight movement directions, or one attack action per alive enemy unit. 
We apply verb-level action masking so that movement is available only when friendly units are alive, and attack is available only when enemies are present. 
This reduces infeasible exploration while preserving centralized multi-unit control.

\textbf{Reward Function.}
The reward combines navigation, combat, and terminal feedback:
\( r_t = r^{\text{nav}}_t + r^{\text{combat}}_t + r^{\text{terminal}}_t \).
Navigation reward is based on the change in average distance from living friendly units to the beacon, giving positive reward for approaching and negative reward for retreating. 
Combat reward combines distance progress toward the enemy centroid, health-based shaping for damage dealt and received, and event-based bonuses or penalties for kills and casualties.

\textbf{Terminal Outcome.}
Terminal rewards are outcome dependent: $+25$ for beacon capture, $+10$ for combat victory, $-10$ for combat loss, and $-15$ for timeout. 
Ties receive no additional terminal reward.

\input{figure/Appendix/5_Exp_2_and_3_Results}
%%%%%%%%%%%%%%%%%%%%%%%%%%%%%%%%%%%%%%%%%%
\section{Pilot 3: Camera Lock}
\label{appendix:pilot3}

Building on Pilot 2, we keep the observation representation and reward function unchanged while refining two design choices: camera locking and stronger action masking.

\begin{wrapfigure}{r}{0.48\textwidth}
    \vspace{-1.2em}
    \centering
    \includegraphics[width=\linewidth]{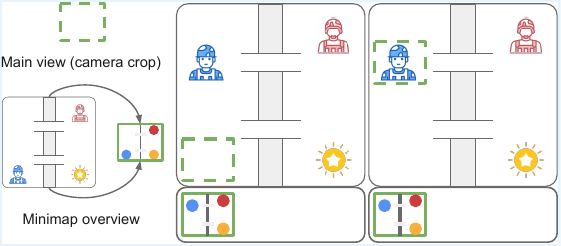}
    \caption{
    Camera-lock observation setup in Pilot 3.
    \textbf{Left:} legend for screen and minimap views.
    \textbf{Middle:} under a free camera, the screen does not track unit movement.
    \textbf{Right:} with camera lock, the screen remains centered on the friendly unit group.
    Camera locking keeps relevant units and nearby terrain visible without adding camera-control actions.
    }
    \label{fig:camera_lock}
    \vspace{-1.0em}
\end{wrapfigure}

\textbf{Camera Lock.}
In SC2, camera movement is an explicit action. 
With a free camera, units can move out of the screen view unless the agent also learns camera-control behavior, forcing it to reason about both \emph{where to look} and \emph{what to do}. 
To remove this confound, we lock the camera to the friendly unit group at every timestep. 
This keeps allied units, nearby terrain, and proximal enemies visible in the screen observation, while the minimap still provides compact global context. 
Thus, the agent receives more informative visual input without expanding the action space.

\textbf{Action Space.}
Pilot 3 extends the structured action interface from Pilot 2 with branch-level action masking over the \texttt{verb}, \texttt{who}, \texttt{direction}, and \texttt{enemy\_idx} branches. 
The \texttt{who} mask permits only alive friendly units, \texttt{enemy\_idx} permits only alive enemy targets plus a null target, and \texttt{direction} enables movement directions only when movement is feasible. 
The \texttt{verb} mask permits \emph{move} and \emph{attack} only when valid entities are present. 
This reduces invalid joint actions and focuses exploration on executable commands.

\subsection{Training Protocol and Compute Budget}
Pilots~\ref{appendix:pilot2} and~\ref{appendix:pilot3} use the same training protocol. 
We train Maskable PPO~\cite{huang2020closer} using the SB3 implementation with default hyperparameters and no algorithm-specific tuning. 
Each environment variant is trained for 5M timesteps. 
Across all evaluated difficulty levels and task variants, this yields 18 independent training runs, with each run requiring approximately 5--6 days on a single consumer-grade GPU.

%%%%%%%%%%%%%%%%%%%%%%%%%%%%%%%%%%%%%%%%%%
\section{Results and Qualitative Analysis}

Figure~\ref{fig:exp2and3_results} summarizes the terminal outcome distributions for Pilots~2 and~3. 
While these results provide a coarse view of performance, qualitative inspection shows that many successes arise from degenerate or incidental behaviors rather than reliable objective reasoning.

\subsection{Pilot 2: Qualitative Behavior}

In \textbf{V1}, the agent learns a combat-dominant strategy, typically sending the full friendly group toward enemy engagement. 
Combat wins are frequent due to numerical advantage, while navigation wins occur only when the beacon lies along the combat path.

In \textbf{V2}, the agent often begins with full-force combat and then retreats along a fixed upper-right trajectory after partial unit loss. 
Navigation wins occur only when the beacon lies on this retreat path; otherwise, the agent frequently times out. 
Combat outcomes remain stochastic due to balanced firepower rather than adaptive tactics.

In \textbf{V3}, performance drops sharply. 
In Base and Combat layouts, the agent initiates unfavorable engagements and often fragments its forces, leading to combat losses. 
In Navigate layouts, it performs limited region-specific exploration before idling, with rare navigation wins only when the beacon is incidentally encountered.

\subsection{Pilot 3: Qualitative Behavior}

With camera locking, the agent generally selects all friendly units and issues group-level movement or attack commands. 
However, a camera-centered failure mode emerges: once visible enemies are eliminated, enemies outside the camera view are often ignored, causing repeated no-op actions and timeout losses.

In \textbf{V1}, behavior remains combat-dominant across layouts, with little evidence of navigation targeting or tactical coordination. 
In \textbf{V2}, behavior varies: the agent delays before combat in Base, commits directly to combat in Combat, and oscillates near bridge regions in Navigate, often failing to commit to either objective. 
In \textbf{V3}, the agent either rushes into poor combat engagements or follows fixed edge trajectories, sometimes avoiding enemies and sometimes missing the beacon depending on its placement.

Overall, these Pilot 2 and Pilot 3 results were later found to be unreliable because the attack-action interface introduced a data leak: the availability of enemy-specific attack IDs implicitly revealed enemy presence and position. 
This shortcut weakened the intended search problem, allowing the policy to infer objective-relevant information without genuinely exploring the map.

%% file: figure/Appendix/5_Exp_2_and_3_Results.tex
\begin{figure*}[t]
    \centering
    % -------- Row 1 --------
    \begin{subfigure}{0.31\textwidth}
        \centering
        \includegraphics[width=\linewidth]{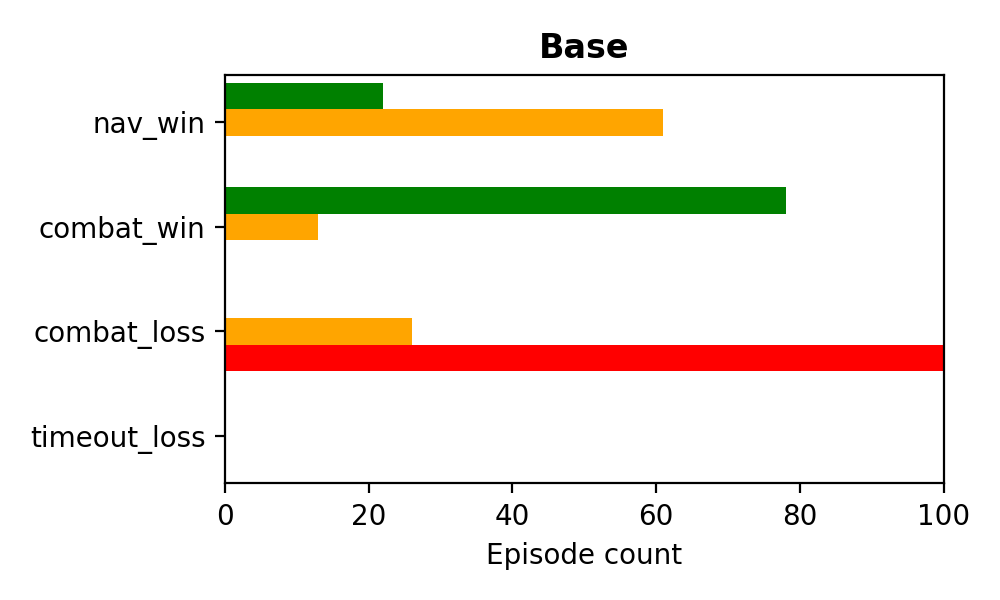}
    \end{subfigure}\hfill
    \begin{subfigure}{0.31\textwidth}
        \centering
        \includegraphics[width=\linewidth]{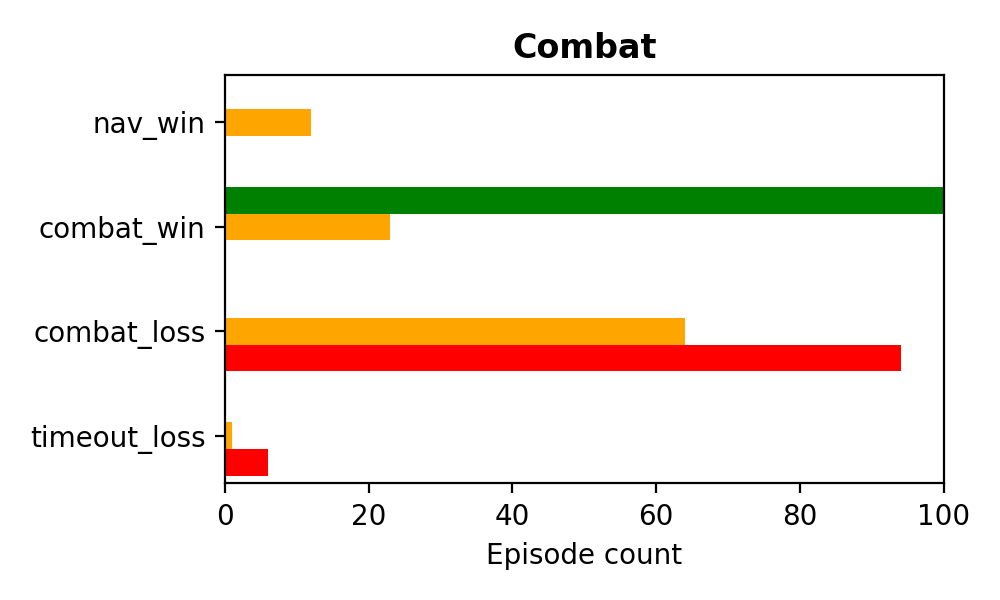}
    \end{subfigure}\hfill
    \begin{subfigure}{0.31\textwidth}
        \centering
        \includegraphics[width=\linewidth]{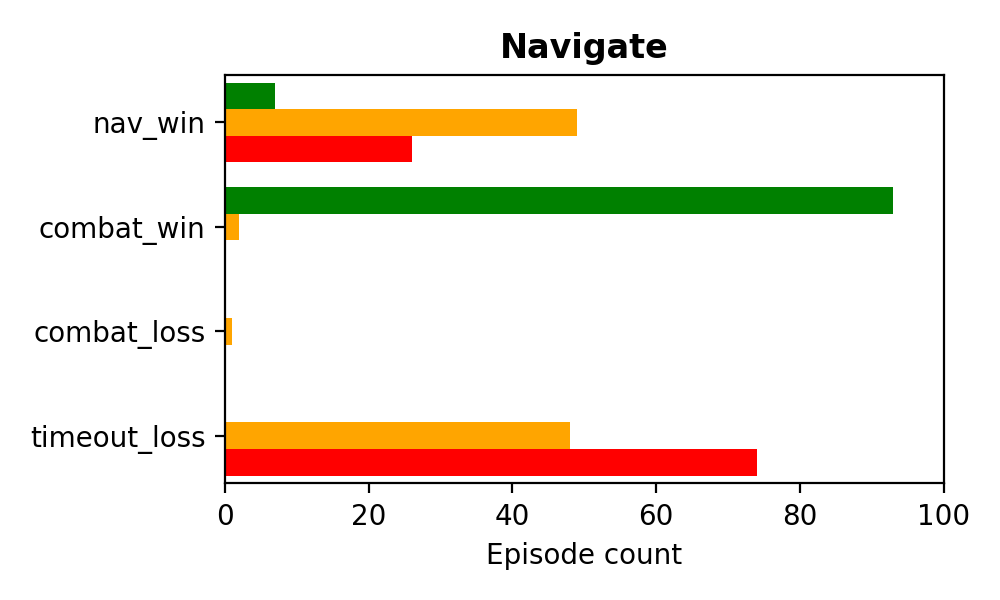}
    \end{subfigure}
    % -------- Row 2 --------
    \begin{subfigure}{0.31\textwidth}
        \centering
        \includegraphics[width=\linewidth]{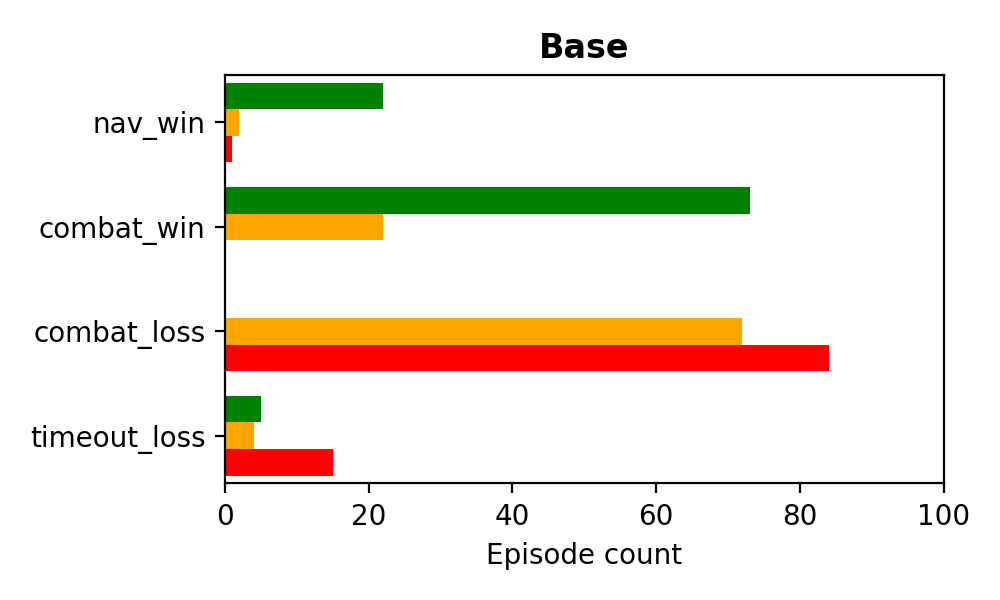}
    \end{subfigure}\hfill
    \begin{subfigure}{0.31\textwidth}
        \centering
        \includegraphics[width=\linewidth]{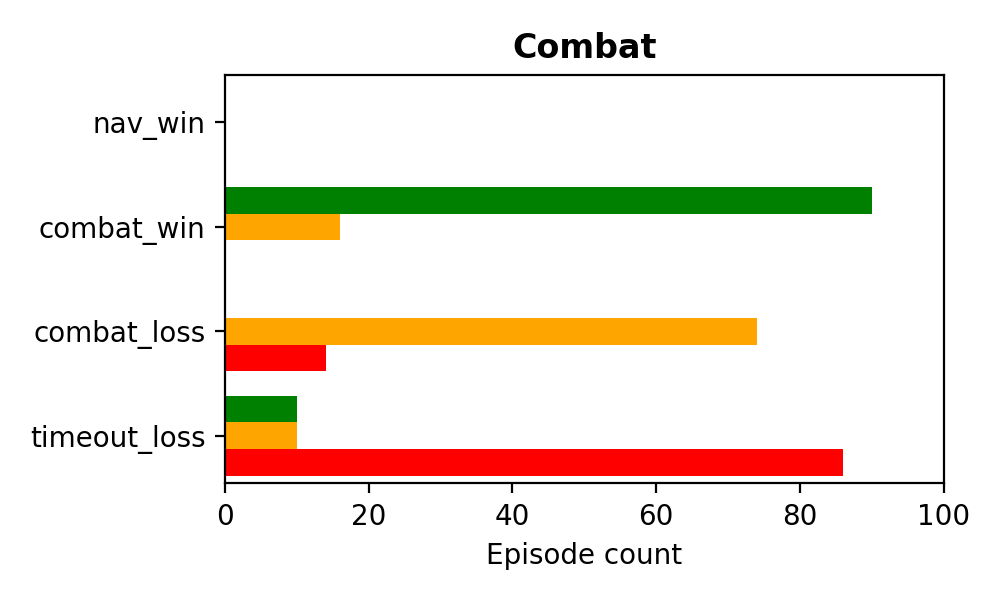}
    \end{subfigure}\hfill
    \begin{subfigure}{0.31\textwidth}
        \centering
        \includegraphics[width=\linewidth]{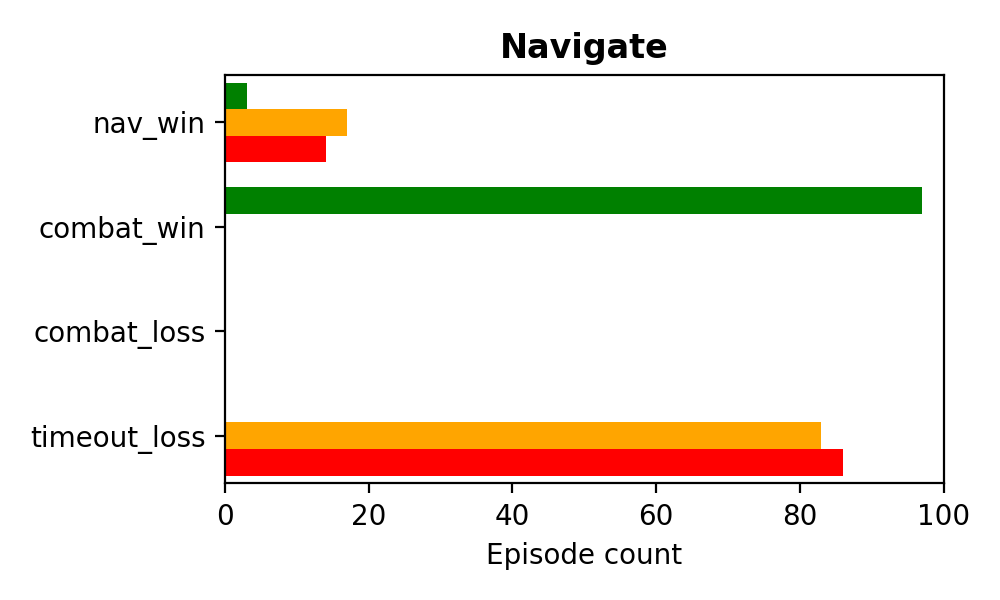}
    \end{subfigure}
    \caption{
    Terminal outcome distributions for \textbf{Experiment 2 (top)} and \textbf{Experiment 3 (bottom)} across all variants.
    Each plot corresponds to a variant layout and aggregates outcomes across all three unit-count settings. Colors indicate map difficulty: V1 (green), V2 (orange), and V3 (red).
    }
    \label{fig:exp2and3_results}
\end{figure*}

%% file: documentBody/Appendix/D-Trigger.tex
\section{Trigger Construction Details}\label{triggers}
This section documents the trigger-level construction of the Two-Bridge Map Suite using the SC2 Editor.
All map files are provided in the supplementary material.
To inspect the trigger logic, users should first install SC2 and copy the provided map files into the appropriate \texttt{StarCraft II/Maps} directory. 
Opening any of these maps launches the StarCraft~II Editor. 
Within the editor, clicking the Triggers icon (represented by a gears symbol in the editor toolbar) opens the trigger editor, where the trigger configurations are described.
The figures in this section illustrate the trigger logic for the \texttt{V2} variants, in which both players are initialized with equal unit counts. 
The triggers specify (i) randomized spawn region selection, (ii) beacon placement logic, and (iii) unit creation and camera initialization. 
Different unit-count variants can be generated by modifying the player unit variables (e.g., \texttt{P1Units}, \texttt{P2Units}) within the same trigger framework, without altering the overall map structure.

Importantly, episode termination conditions and game-control logic are \emph{not} implemented in the SC2 triggers.
All termination criteria (e.g., win, loss, timeout) are handled exclusively within the Gym-compatible environment wrappers.
This separation avoids trigger-level side effects and potential in-game control glitches, and ensures consistent episode semantics across layouts and algorithms.

\input{figure/Appendix/B1_Base}
\textbf{Base Layout Trigger Logic:}
Figure~\ref{fig:base} shows the SC2 Editor trigger configuration for the Base layout. 
At map initialization, the spawn region for Player~1 is first selected uniformly at random from regions \texttt{R1–R3}. 
A fixed number of units (five in the V2 configuration) are then spawned for Player~1 at random points within the selected region, and the camera is panned to center on this spawn location.
Next, the beacon spawn region is selected uniformly at random from the regions \texttt{R4–R6}, and the beacon is instantiated at a random point within the chosen region.
The spawn region for Player~2 is then selected from the remaining regions not occupied by the beacon, ensuring that player and objective placements do not overlap. 
Player~2 units are spawned analogously within the selected region.
Finally, fog-of-war is disabled to provide full observability.

\input{figure/Appendix/B2_Combat}
\textbf{Combat Layout Trigger Logic:}
Figure~\ref{fig:combat} shows the SC2 Editor trigger configuration for the Combat layout. 
At map initialization, the beacon spawn region is first selected uniformly at random from regions \texttt{R1–R3} and instantiated at a random point within the chosen region.
Next, the spawn region for Player~1 is selected uniformly at random from regions \texttt{R4–R6}, and a fixed number of units are spawned at random points within this region. 
The camera is then panned to the Player~1 spawn location.
The spawn region for Player~2 is subsequently selected from the remaining regions not occupied by Player~1, ensuring spatial separation between the two players. 
Player~2 units are spawned analogously within the selected region.
Finally, fog-of-war is disabled to provide full observability.

\input{figure/Appendix/B3_Navigate}
\textbf{Navigate Layout Trigger Logic:}
Figure~\ref{fig:navigate} shows the SC2 Editor trigger configuration for the Navigate layout. 
At map initialization, the spawn region for Player~2 is first selected uniformly at random from regions \texttt{R1–R3}, and a fixed number of units are spawned at random points within the selected region.
Next, the spawn region for Player~1 is selected uniformly at random from regions \texttt{R4–R6}. 
Player~1 units are then spawned within this region, and the camera is panned to the Player~1 spawn location.
The beacon spawn region is subsequently selected from the remaining regions not occupied by Player~1, ensuring that the navigation objective is spatially decoupled from Player~1’s initial position. 
The beacon is instantiated at a random point within the chosen region.
Finally, fog-of-war is disabled to provide full observability.

\input{figure/Appendix/B4_Camera_Lock}
\textbf{Camera Lock.}
In the camera-lock setting, we retain the same randomized spawn logic for all entities as in the corresponding layout variants.
A camera-lock trigger is additionally applied at map initialization to fix the player camera to the center of Player~1’s units.
Figure~\ref{fig:camLock} illustrates the trigger configuration used to enforce this constraint.

%% file: figure/Appendix/B1_Base.tex
\begin{figure}[h]
    \centering
    \includegraphics[width=\textwidth]{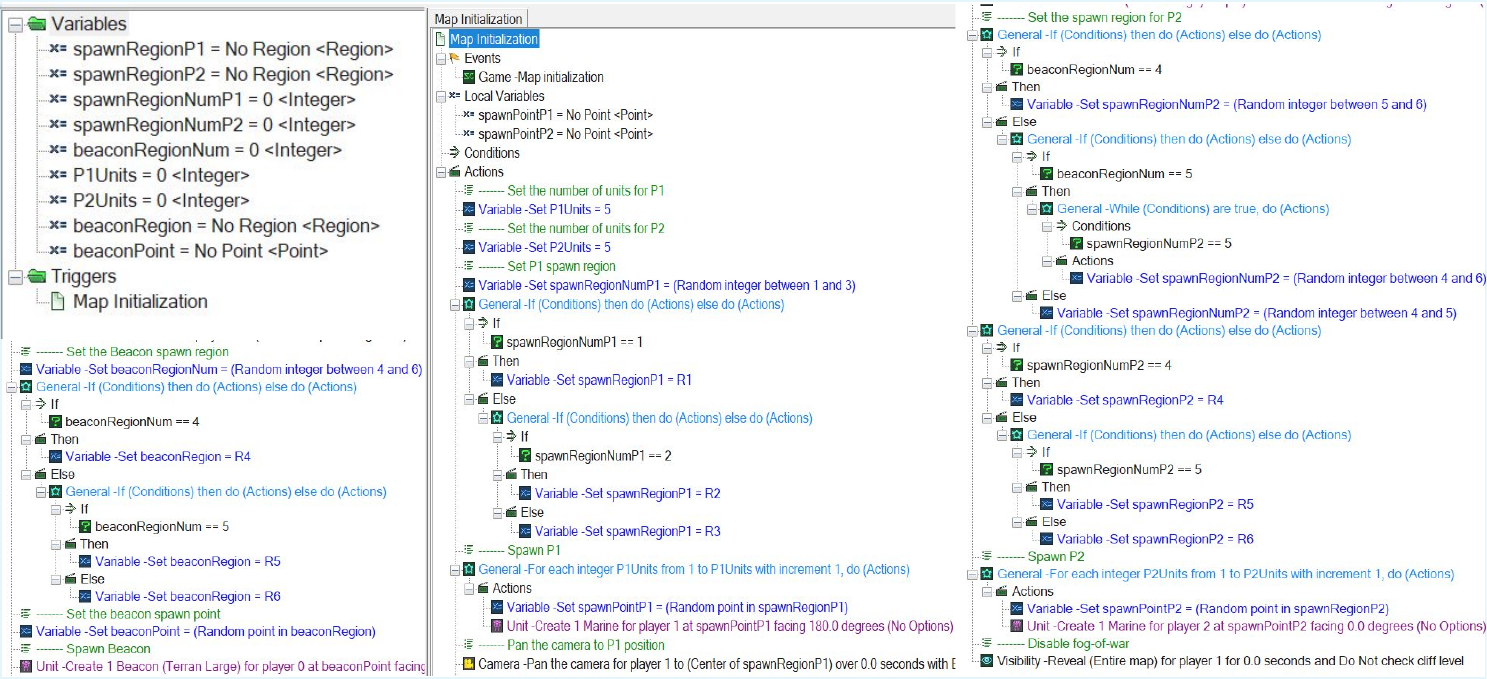}
    \caption{Trigger configuration for the Base layout.}
    \label{fig:base}
\end{figure}

%% file: figure/Appendix/B2_Combat.tex
\begin{figure}[h]
    \centering
    \includegraphics[width=\textwidth]{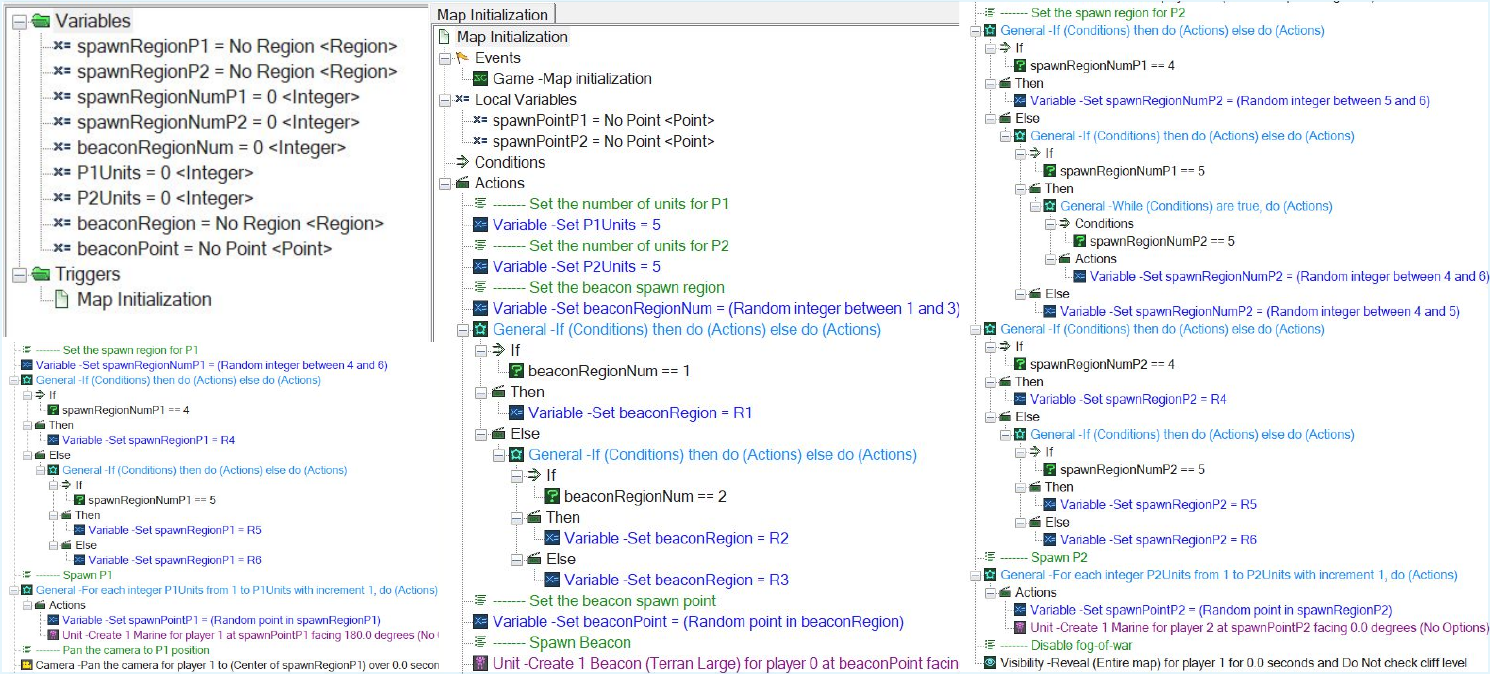}
    \caption{Trigger configuration for the Combat Layout.}
    \label{fig:combat}
\end{figure}

%% file: figure/Appendix/B3_Navigate.tex
\begin{figure}[h]
    \centering
    \includegraphics[width=\textwidth]{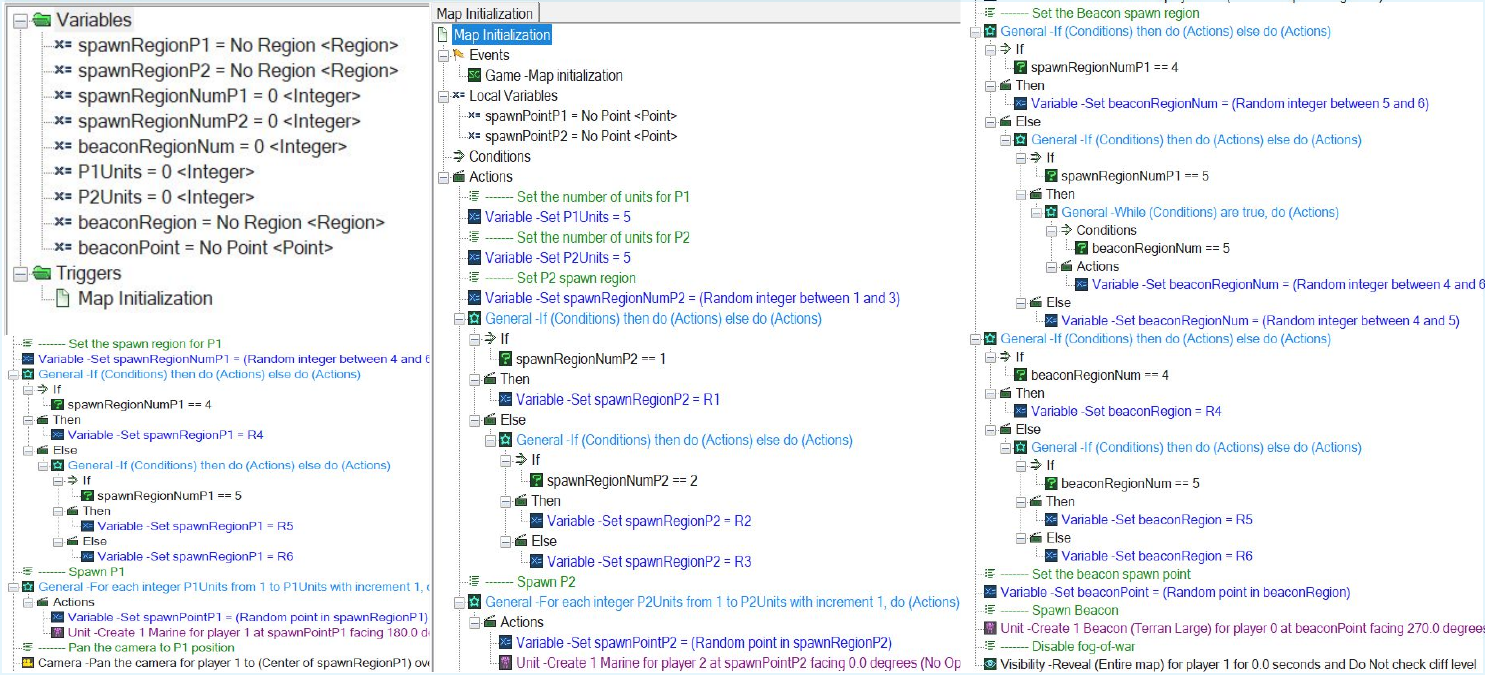}
    \caption{Trigger configuration for the Navigate Layout.}
    \label{fig:navigate}
\end{figure}

%% file: figure/Appendix/B4_Camera_Lock.tex
\begin{figure}[h]
    \centering
    \includegraphics[width=\textwidth]{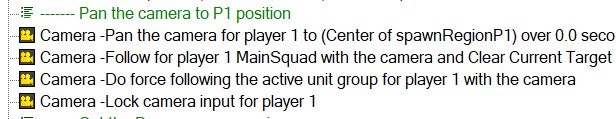}
    \caption{Camera Lock}
    \label{fig:camLock}
\end{figure}